\documentclass[a4paper,fleqn]{cas-dc}

\usepackage{graphicx} 
\usepackage[utf8]{inputenc}
\usepackage[T1]{fontenc}
\usepackage{graphicx}
\usepackage{amsmath}
\usepackage{amsfonts}
\usepackage{amssymb}
\usepackage{url}
\usepackage{ulem} 
\usepackage{xcolor}
\usepackage{physics}
\usepackage{bbm}
\usepackage{bbold}
\usepackage{booktabs}
\usepackage{subcaption}
\usepackage{fvextra}
\usepackage{longtable}
\usepackage{multirow}
\usepackage{tabularx}
\usepackage{array}
\usepackage{hyperref}
\usepackage{lineno}

\usepackage[numbers]{natbib}

\DefineVerbatimEnvironment{prompt}{Verbatim}{
  breaklines=true,
  breakanywhere=true,
  fontsize=\small,
  commandchars=\\\{\}, 
}

\begin{document}
\let\WriteBookmarks\relax
\def\floatpagepagefraction{1}
\def\textpagefraction{.001}
\shorttitle{Tree Classification from HSI/LiDAR}
\shortauthors{M. Romaszewski et~al.}

\title [mode = title]{From Articles to Canopies: Knowledge-Driven Pseudo-Labelling for Tree Species Classification using LLM Experts}
\affiliation[1]{organization={Institute of Theoretical and Applied Informatics, Polish Academy of Sciences},
                addressline={Bałtycka 5}, 
                city={Gliwice},
                postcode={44-100}, 
                country={Poland}}
\affiliation[2]{organization={University of Lodz, Faculty of Biology and Environmental Protection, Department of Biogeography, Paleoecology and Nature Conservation},
                addressline={Banacha 1/3}, 
                postcode={90-237}, 
                city={Lodz},
                country={Poland}}
\affiliation[3]{organization={MGGP Aero},
                addressline={Kaczkowskiego 6}, 
                postcode={33-100}, 
                city={Tarnów},
                country={Poland}}
\affiliation[4]{organization={University of Warsaw, Faculty of Geography and Regional Studies, Department of Geoinformatics, Cartography and Remote Sensing},
                addressline={Krakowskie Przedmieście 30}, 
                postcode={00-927}, 
                city={Warszawa},
                country={Poland}}    

\author[1]{Michał Romaszewski}[orcid=0000-0002-8227-929X]
\ead{mromaszewski@iitis.pl}
\author[2,3]{Dominik Kopeć}[orcid=0000-0003-0831-2992]
\author[1]{Michał Cholewa}[orcid=0000-0001-6549-1590]
\author[1]{Katarzyna Kołodziej}[orcid=0000-0002-2329-107X]
\author[1]{Przemysław Głomb}[orcid=0000-0002-0215-4674]
\author[3]{Jan Niedzielko}[orcid=0000-0002-6531-0267]
\author[3]{Jakub Charyton}[orcid=0000-0003-2052-9051]
\author[3]{Justyna Wylazłowska}[orcid=0000-0001-6015-0463]
\author[4]{Anna Jarocińska}[orcid=0000-0003-3610-4656]

\begin{abstract}
Hyperspectral tree species classification is challenging due to limited and imbalanced class labels, spectral mixing (overlapping light signatures from multiple species), and ecological heterogeneity (variability among ecological systems). Addressing these challenges requires methods that integrate biological and structural characteristics of vegetation, such as canopy architecture and interspecific interactions, rather than relying solely on spectral signatures. This paper presents a biologically informed, semi-supervised deep learning method that integrates multi-sensor Earth observation data, specifically hyperspectral imaging (HSI) and airborne laser scanning (ALS), with expert, ecological knowledge. The approach relies on biologically inspired pseudo-labelling over a precomputed canopy graph, yielding improved classification with a lightweight classifier. In addition, ecological priors are constructed using an LLM-assisted, citation-constrained workflow and encoded as a cohabitation matrix with likelihoods of species occurring together. These priors are incorporated into the pseudo-labelling strategy, effectively introducing expert knowledge into the model. Experiments on two real-world forest datasets showed macro-F1 improvements of 6.01 and 4.85 percentage points over the strongest classical HSI--ALS reference pipelines.
\end{abstract}

\begin{keywords}
Hyperspectral \sep ALS \sep LLM \sep Remote Sensing \sep Deep Learning
\end{keywords}
\maketitle
\section{Introduction}
The world's forests are critical habitats for biodiversity~\cite{Brockerhoff2017biodiversity} and strongly influence climate through complex, planetary scale, physical, chemical, and biological processes~\cite{Bonan2008Climate}. Forests with a naturally diversified mixture of tree species stand out as especially important, due to their enhanced properties of biomass production, soil carbon storage, water yield and soil retention~\cite{Gamfeldt2013multiple,Bai2024diversity}. Consequently, reliable tree species classification and mapping have become foundational inputs for a number of tasks, e.g. biodiversity indices derivation and conservation planning~\cite{Vanguri2024mapping} or formulation of management practices~\cite{Bai2024diversity}. Detailed forest information, including species composition, is viewed as a priority source for EU-level policy decisions, such as forest threats mitigation~\cite{Atzberger2020monitoring} or accurate planning and management of forests~\cite{EUSTAFOR2024position}. In parallel, rapid advances in Earth observation and machine learning are making large-scale, species-resolved forest maps technically feasible, from benchmark datasets~\cite{Ahlswede2023dataset} to national tree species products~\cite{Blickensdoerfer2024national}. From a methodological perspective, tree species classification is a demanding testbed for machine learning, as it is a difficult, multiclass problem with sparse and imbalanced labels~\cite{Qin2025multi} that requires accounting for a large variety of ecological conditions and the need for combining biological and machine learning approaches for understanding and quantifying the drivers for classification errors~\cite{Fassnacht2016Review}. Importantly, recent reviews~\cite{Zhong2024Review, Pu2021Mapping} show that most existing approaches treat tree species classification as a purely data-driven pattern recognition task, with limited integration of explicit biological and ecological knowledge about species traits, interactions and habitat constraints. Addressing this gap, we present a tree species classifier that couples dedicated machine learning techniques with explicit biological background knowledge, aiming for more accurate, robust and ecologically interpretable species maps.

Classification of tree species is challenging due to high intra-species variability, which occurs both among individual trees and within a single tree crown. This variability arises because trees of the same species may naturally differ in their biophysical, morphological, and structural characteristics, which, in turn, affect the sensor-derived features observed in remote sensing data~\cite{marconi2022continental, zhang2020extraction}. Spectral reflectance varies depending on the tree species, but this is only one of many factors affecting reflectance values~\cite{Pu2021Mapping}. Essential are biophysical variables of each plant, such as leaf biochemistry (like pigment content), morphology (leaf area), and canopy structure~\cite{marconi2022continental}. Also, the health condition and the generative phase of the trees (appearance of flowers and fruits) influence the reflectance values, especially in the visible and near-infrared range. The size and structure of the tree (including age-related factors) may influence the reflection values recorded by Airborne Laser Scanning (ALS)~\cite{Fassnacht2016Review, Pu2021Mapping, niedzielko_airborne_2024}. Variability within the tree crown is also important, as pixels dominated by reflectance from leaves or branches can be identified through differences in reflectance in the optical range, as well as in ALS data. If a pixel occurs at the edge of the tree, the reflectance may be influenced by the background, including other adjacent plant species~\cite{zhang2020extraction, zhong2022identification}. However, reflectance values may be similar across species due to comparable biophysical plant parameters, as well as blurring of differences under, for example, deteriorating conditions~\cite{shi2021mapping, marconi2022continental}. These factors increase within-species spectral variability and can also cause spectral overlap between different species, making tree species classification a hard problem.

Various approaches are used to identify species, including a combination of different types of data, such as spectral imagery and LiDAR data, which allows the use of biophysical and morphological characteristics of vegetation~\cite{Fassnacht2016Review, Pu2021Mapping}. Another solution is to use multi-temporal data~\cite{damico_data_2024}. A prominent baseline approach for exploring biologically grounded features in tree species classification involves using ALS and hyperspectral imaging (HSI) data, often for computing vegetation indices, along with classical, explainable machine learning (ML) models such as the random forest (RF) classifier~\cite{scholl2020integrating, quan2023tree, kluczek2023mountain}. Segmentation techniques, often used for grouping pixels of individual trees, are also employed~\cite{farsad2022integrating, quan2023tree}. Some works augment data with publicly available Sentinel-2 satellite images~\cite{farsad2022integrating, kluczek2023mountain}. 

Artificial neural networks, especially deep learning (DL) models~\cite{graves_data_2023}, often outperform other methods in classification from remote sensing images by extracting hierarchical features from complex spectral, spatial, and structural data~\cite{guo_individual_2022, wang_hybrid_2024}. Recently, a semi-supervised DL approach has become prominent for pixel-level tree species; for example, Ni-Meister et al. \cite{ni2024assessing} used SwAV-based pretraining on unlabelled hyperspectral data followed by supervised fine-tuning. Such deep learning approaches can outperform classical ML models, especially when sufficient training data or effective pretraining is available.

However, accurate species classification remains challenging. Both traditional methods and DL approaches work well for the most common classes. However, both struggle with imbalanced data and rare species~\cite{graves_data_2023}. Performance reported in tree species identification competition~\cite{graves_data_2023} rarely exceeded~$\sim0.32$ F1.

Semi-supervised models are designed for settings where labelled data are scarce but unlabelled data are abundant, combining a small labelled set with a much larger unlabelled set to learn more robust decision boundaries and improve generalization. For example, the approach in~\cite{li2024pseudo} employs contrastive learning to address the label scarcity problem and reduce the semantic gap between multimodal features; the method combines unsupervised feature extraction with supervised classification. \cite{wang2023nearest} introduces the Nearest Neighbor-based Contrastive Learning Network (NNCNet), which integrates a nearest-neighbor-based data augmentation in which the nearest neighbors of positive samples are fed into the encoder. \cite{lu2023coupled} uses coupled adversarial feature learning (CAFL) and multi-level feature fusion classification (MFFC). \cite{guo2023semi} uses unsupervised cross-domain feature fusion with spectral attention, where unlabelled sub-images are used to conduct feature fusion pre-training and a few available labelled samples are used for feature optimisation. Label scarcity~\cite{wang2023nearest} is a typical and significant problem for such algorithms.

Because of that, recent HSI classification research has increasingly moved toward methods that reduce the need for dense pixel-level annotations or improve transferability across datasets. For example, ITER explores weakly supervised image-to-pixel learning from image-level tags \cite{yang2024iter}, while HyperFree introduces a channel-adaptive, tuning-free hyperspectral foundation model for transferable HSI interpretation \cite{li2025hyperfree}.
These approaches represent complementary directions to the method proposed here. ITER reduces annotation requirements through weak supervision, while HyperFree addresses transferability through large-scale pre-training and a foundation-model paradigm. Our framework instead focuses on incorporating explicit, domain-specific ecological knowledge into the learning process. It uses a lightweight classifier and an interpretable cohabitation prior to constrain pseudo-label assignment, without requiring large-scale pre-training. Thus, rather than competing directly with weakly supervised or foundation-model approaches, the proposed method addresses a different aspect of limited-label HSI classification: the explicit use of ecological constraints that are difficult to derive from spectral data alone.

Additionally, recent work has increasingly emphasized pseudo-labelling --- a strategy in which the model's high-confidence predictions on unlabelled data are used to extend the labelled training set in subsequent passes --- as a central mechanism in semi-supervised classification frameworks~\cite{farsad2022integrating, hou2025integrating}. 
A typical criterion for pseudo-labelling of samples is spectral similarity, often accompanied by spatial relationship, according to which neighbouring pixels in a homogenous region are highly probable to belong to the same class~\cite{wang2023advances}. However, several studies~\cite{zhong2022identification, zhang2020extraction} highlight that only spectral signatures belonging to the centroid of the individual tree provide characteristics useful for tree identification. In contrast, at canopy edges, presence of mixed pixels leads to misclassification.  In related tree species classification studies~\cite{ni2024assessing}, training sample preparation is restricted to crown centres or treetop locations in order to reduce spectral mixing and obtain higher-quality spectral information. However, such conservative sample selection further limits the number of available training samples. Pseudo-labelling addresses this limitation by adding high-confidence predictions for unlabelled samples to the training set in subsequent training stages~\cite{farsad2022integrating,hou2025integrating}.

The fundamental bottleneck behind limited labelled samples is that tree species identification relies on field surveys, a process that is labour-intensive, costly, and constrained by limited throughput, the need for expert knowledge, and the logistical complexity of accessing remote or difficult terrain~\cite{tong2024tree, shi2021mapping, jiang2023seeing}. While the increasing availability of unmanned aerial vehicles (UAVs) has substantially improved access to large-scale, high-resolution HSI and ALS data, enabling comprehensive coverage of entire areas of interest \cite{mayra2021tree}, labelled samples are still scarce. While small, such datasets are valuable, as they often provide unique insights into rare, endangered species and ecological processes~\cite{hirn2024transfer, todman2023small}. However, such datasets are typically class-imbalanced and insufficient for training DL models, which tend to overfit and, as a result, exhibit poor test performance under severely limited supervision \cite{zhang2023features}. For example,~\cite{tong2022spectral} showed that simulated restriction of the training samples' number drastically degraded CNN-based classifier accuracy on the tree species classification task. Simultaneously, the authors demonstrated that architectures based on hierarchical traditional ML approaches remained much more robust in terms of accuracy under such restrictions. This clearly suggests that well-adapted ML approaches are a promising alternative to CNN-based approaches for the sparse-sample HSI classification task. Another requirement is to consider the crown-edge mixed-pixel (spectral mixing) effect, where boundary pixels contain a mixture of canopy and background reflectance, reducing pixel purity and increasing misclassification~\cite{zhong2022identification, zhang2020extraction}. However, with high-resolution HSI and ALS, individual tree canopies can often be delineated directly from unlabelled data by exploiting biologically informed morphological and structural cues~\cite{ni2024assessing, quan2023tree}. This suggests a reliable approach that couples unsupervised individual tree crown delineation with a robust classifier to label unlabelled canopy tops based on their proximity to labelled trees.

This biologically inspired strategy is well-motivated, but it overlooks a key ecological factor: the local co-occurrence of tree species within the same area, which could provide valuable contextual information for classification. Deriving cohabitation (co-occurrence) information is not straightforward, because it is typically reported in the literature in heterogeneous, non-standardized forms, making it difficult to compile into a consistent, usable dataset. In~\cite{farsad2022integrating}, the expert system is built to establish the rules determining the probabilities of vegetation classes' occurrence dependent on propagation direction. However, the rules are manually created from a large set of biological metadata and field surveys strictly related to the study area, which incurs substantial costs and workload and limits scalability. To mitigate this issue, \cite{hirn2024transfer} proposes a transfer learning framework that uses a model trained on a large dataset of plant species cohabitation patterns to infer plant relations in much smaller datasets from different regions. This proposal, although much more scalable, requires a large pre-training dataset, which needs to be obtained through extensive and costly field research. 

In light of these limitations, recent advances in generative, cloud-based artificial intelligence offer a practical way to reduce the dependency on expensive field-derived co-occurrence datasets. State-of-the-art large language models (LLMs) are increasingly effective at retrieving, synthesizing, and interpreting publicly available scientific information, providing a scalable mechanism to extract domain knowledge from the literature for downstream use~\cite{gougherty2024testing}. They are also becoming capable assistants for quantitative scientific analyses~\cite{hayes2025conversing}, which opens the possibility of automatically converting heterogeneous ecological descriptions into structured priors. Building on this capability, we turn to the scientific literature as a rich, yet currently underutilized, source of cohabitation evidence. Specifically, we propose to screen publications for co-occurrence statements among a predefined set of species labels and automatically translate these into a species cohabitation matrix. This matrix can then be used to inject, automatically, current expert ecological knowledge into our biologically motivated neighborhood-based pseudo-labelling framework, providing a scalable alternative to manual rule construction or extensive field campaigns.

Specifically, this paper makes the following contributions to tree species classification with sparse and imbalanced field reference data:

\begin{itemize}
\item \textit{We propose a semi-supervised HSI--ALS framework that extends scarce tree-level annotations with biologically informed pseudo-labels.}
The method combines a lightweight dual-stream neural classifier with a canopy graph derived from treetop detections. Rather than attempting to label the entire scene automatically, it adds selected high-confidence treetop observations located near field-verified trees to the training set. This addresses a practical limitation of tree species mapping: reliable botanical annotations are costly to obtain, whereas unlabelled canopy observations are abundant. On the Wigierski dataset, the resulting DSNN+P model achieved 79.38 macro-F1, improving by 6.01 percentage points over the strongest classical HSI--ALS reference pipeline and by 1.96 percentage points over the non-pseudo-labelled DSNN. On the Bory Tucholskie dataset, DSNN+P achieved 77.04 macro-F1, improving by 4.85 percentage points over the strongest classical HSI--ALS reference pipeline and by 3.73 percentage points over the non-pseudo-labelled DSNN.

\item \textit{We introduce an LLM-assisted procedure to construct an ecological cohabitation prior for pseudo-label assignment.}
The LLM is used to extract and organise ecological evidence from literature, habitat information, and regional forest knowledge into a species cohabitation matrix. This matrix acts as a soft plausibility prior that complements classifier confidence and spatial proximity during pseudo-label selection. Expert assessment indicated that the generated matrix was broadly consistent with regional forest ecology. The proposed procedure, therefore, makes ecological background knowledge explicit, inspectable, and usable within the training-set expansion process.

\item \textit{We analyse when knowledge-guided pseudo-labelling is useful and how its components contribute to performance.}
The approach is intended for forests where high-quality field labels are limited, but unlabelled HSI--ALS observations and ecological knowledge are available. The evaluation compares the method with classical HSI and HSI--ALS pipelines, tests it on two forest datasets, and includes variants with and without the cohabitation prior. The results indicate that pseudo-labels are useful primarily as an extension of field data, not as a replacement for verified annotations. They also show that the ecological prior should be interpreted as a plausibility constraint rather than as the sole source of performance gain. This makes the method particularly relevant in protected, heterogeneous, or otherwise data-scarce forests, where harvesting-derived labels are unavailable, and augmentation using synthetic multimodal observations may be difficult to apply.
\end{itemize}

Together, these contributions support a practical form of knowledge-guided tree species mapping. The proposed framework combines field-verified labels, real HSI--ALS observations, canopy structure, classifier confidence, and ecological plausibility to expand the training set in a transparent and conservative way. In this setting, the role of the LLM is not to replace ecological expertise or direct validation, but to help translate dispersed ecological evidence into a structured prior that experts can inspect and use for pseudo-label selection.

\section{Method}

\subsection{Dataset}
The dataset contains co-registered ALS and HSI data acquired over the extended area of Wigierski National Park  in August 2019. The area covered by the data acquisition is 158.3 $km^2$. Data collection was carried out using a Vulcanair P-68 Observer 2 aircraft at an average altitude of approximately 1275 m above ground level, following north–south oriented flight lines designed to ensure full coverage with cross-track overlaps of 50\% for ALS and 30\% for HSI scenes. The HSI data was acquired with a HySpex VS-725 system composed of a VNIR-1800 and two SWIR-384 sensors, covering 400–2500 nm with ground sampling distances of 1 m in both the visible–near-infrared and shortwave infrared ranges, while ALS data was collected with a Riegl VQ-780i laser scanner operating at 1064 nm with a scan angle of 50° and a point density of approximately 7.5 points per m$^2$. HSI and ALS data were collected synchronously during three flight campaigns: 17, 24, and 25 of August 2019 under clear-sky conditions with a minimum solar elevation of 30°. 

HSI and ALS data underwent a multi-stage processing workflow to derive geometrically and radiometrically consistent products suitable for quantitative analysis. The ALS point cloud was generated, calibrated, and georeferenced, cleaned from extreme elevation outliers, and classified in TerraScan \cite{TerraScan} into ground, low, medium, and high vegetation, buildings, water, and noise classes, followed by internal visual quality control. HSI was subjected to parametric geometric correction in PARGE \cite{PARGE} using GPS/INS trajectories and ALS-based DSM, resampled with nearest neighbour and referenced to local coordinate system (PL-1992). Atmospheric correction was applied using ATCOR4 \cite{ATCOR4} to convert at-sensor radiance to surface reflectance. The corrected scenes were mosaicked into one hyperspectral cube (1 m spatial resolution, 430 bands, ENVI, BSQ format)

To create labels, field measurements were carried out between June and August 2019. Reference data were acquired using distributed purposive sampling. The sampling design was optimised to cover the spatial and environmental heterogeneity of the study area, ensuring representative coverage of the selected tree species. Because field-based annotation is highly labour-intensive, it produced accurate but sparse reference labels, ultimately covering less than 1 p.p. of the tree population. Reference tree coordinates were captured using a GNSS Trimble DA1 antenna, with a maximum error of 1 m. The recorded coordinates were used as 2D planimetric coordinates $(x,y)$ indicating the horizontal position of the central point of each tree crown. Height information was derived separately from ALS/CHM. Measurements were recorded on mobile devices using an Android-based mobile app. Recorded points were labelled with the tree species name or an indication of tree death, along with a radius value, enabling circular polygons to be constructed from these points. The radius was defined to ensure the polygon did not exceed the actual size of the tree crown; therefore, the rasterised reference polygon contained no pixels outside the tree crown. During post-processing, all GNSS points were visually checked against the orthophotomap and co-registered HSI/ALS products. When a recorded point did not coincide with the visible crown centre, it was shifted to the crown centre before polygon construction and rasterisation.

\begin{figure}
        \centering
        \includegraphics[width=1.0\linewidth]{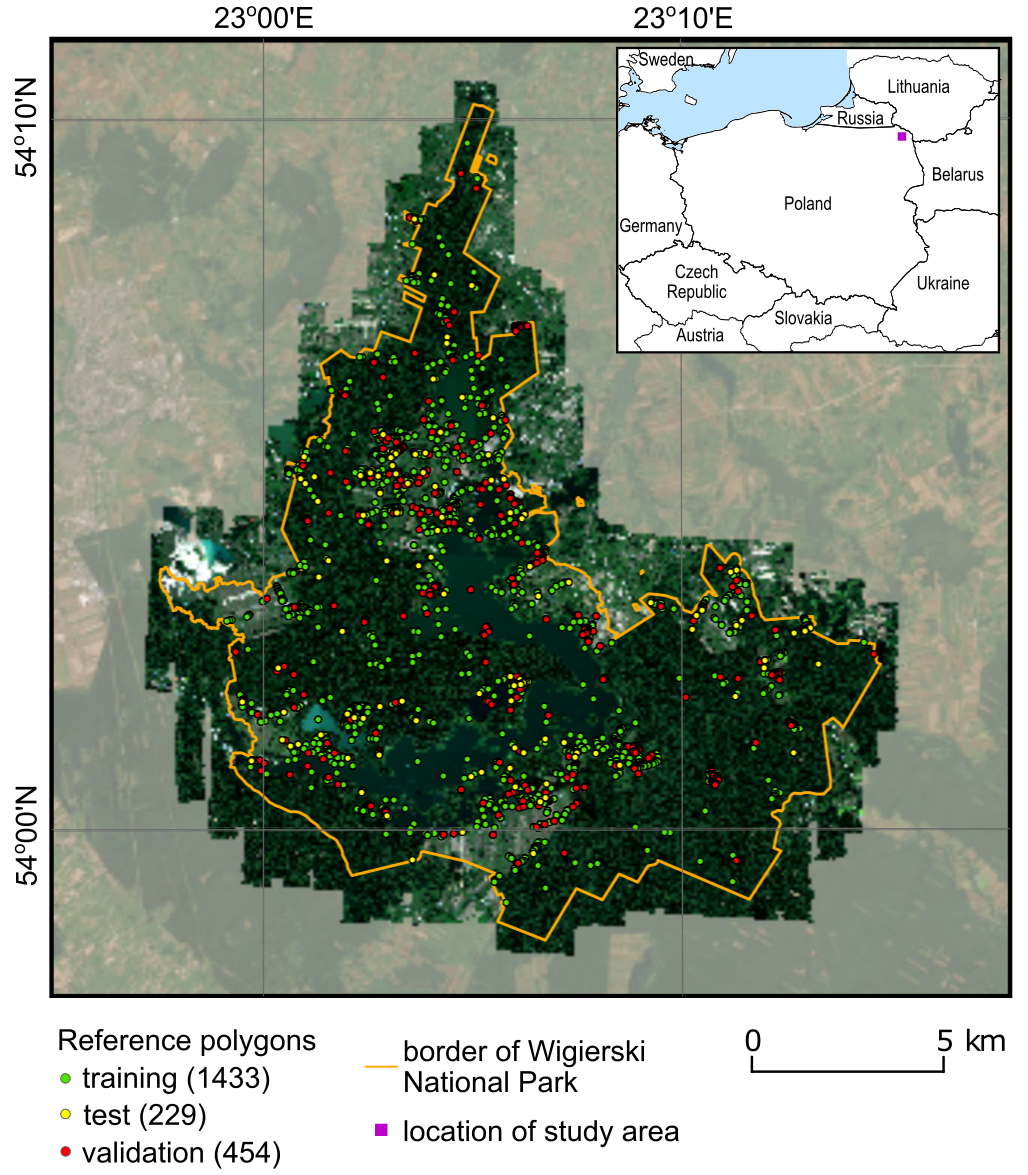}
        \caption{Visualisation of the Wigierski National Park with sampled trees divided into training, validation, and tests subsets}
        \label{fig:study_area}
    \end{figure}

A total of 942 reference trees were included in the analysis: 799 trees assigned to 16 taxa classes and 143 trees labelled as `Other trees' -- comprising less abundant species that could not form a separate taxa class due to insufficient representation in the study area. Reference tree counts by taxa class are presented in Table~\ref{tab:ref_tree_count}. This collection includes all the main forest-forming species found in the study area.  Visualisation of the study area and the training, validation, and test subsets used in the experiments is presented in Figure~\ref{fig:study_area}.

\begin{table*}[ht]
    \centering
    \begin{tabularx}{\textwidth}{l >{\raggedright\arraybackslash}X *{8}{>{\raggedright\arraybackslash}X}}
        \toprule
         &  &
        \multicolumn{2}{c}{Total} & 
        \multicolumn{2}{c}{Training} & 
        \multicolumn{2}{c}{Validation} & 
        \multicolumn{2}{c}{Test} \\
        \cmidrule(lr){3-4} \cmidrule(lr){5-6} \cmidrule(lr){7-8} \cmidrule(lr){9-10}
        Tree taxa class & Class code & Polygons & Pixels & Polygons & Pixels & Polygons & Pixels & Polygons & Pixels \\
        \midrule
        \textit{Acer platanoides} & Ace-pla          & 40  & 1136 & 28 & 758  & 8  & 233 & 4  & 145 \\
        \textit{Alnus glutinosa} & Aln-glu          & 104 & 2298 & 68 & 1640 & 24 & 525 & 12 & 133 \\
        \textit{Betula} spp. & Bet-spp              & 99  & 1304 & 72 & 912  & 21 & 309 & 6  & 83  \\
        \textit{Carpinus betulus} & Car-bet         & 18  & 591  & 10 & 333  & 4  & 162 & 4  & 96  \\
        \textit{Fraxinus excelsior} & Fra-exc       & 26  & 545  & 18 & 365  & 6  & 151 & 2  & 29  \\
        \textit{Larix decidua} & Lar-dec            & 29  & 659  & 22 & 497  & 4  & 122 & 3  & 40  \\
        \textit{Picea abies} & Pic-abi              & 103 & 1646 & 64 & 940  & 26 & 468 & 13 & 238 \\
        \textit{Picea abies} - dead & Pic-dea       & 35  & 486  & 17 & 251  & 13 & 166 & 5  & 69  \\
        \textit{Pinus sylvestris} & Pin-syl         & 101 & 1221 & 73 & 923  & 15 & 141 & 13 & 157 \\
        \textit{Pinus sylvestris} - dead & Pin-dea  & 27  & 245  & 22 & 221  & 3  & 19  & 2  & 5   \\
        \textit{Populus tremula} & Pop-tre          & 31  & 808  & 23 & 566  & 4  & 146 & 4  & 96  \\
        \textit{Quercus robur} & Que-rob             & 54  & 2376 & 35 & 1594 & 10 & 477 & 9  & 305 \\
        \textit{Salix} trees & Sal-tre                & 23  & 889  & 15 & 576  & 6  & 209 & 2  & 104 \\
        \textit{Salix} bushes & Sal-bus              & 40  & 669  & 27 & 451  & 7  & 101 & 6  & 117 \\
        \textit{Tilia cordata} & Til-cor             & 56  & 1885 & 35 & 1024 & 11 & 423 & 10 & 438 \\
        Deciduous trees - dead & Dec-dea            & 13  & 104  & 9  & 71   & 3  & 31  & 1  & 2   \\
        Other trees & Oth-tre                     & 143 & 2383 & 93 & 1670 & 34 & 354 & 16 & 359 \\
        Background & Backgr                     & 1174 & 27489 & 802 & 18014 & 255 & 6838 & 117 & 2637 \\
        \bottomrule
    \end{tabularx}
    \caption{Reference tree count by taxa class (polygons and pixels count across total, training, validation, and test sets)}
    \label{tab:ref_tree_count}
\end{table*}

After post-processing of the field dataset, 1174 background reference polygons were added manually based on the orthophotomap in QGIS. The background class was needed to classify the pixels outside of tree crowns as a background and not as any of tree taxa class. The whole reference dataset was split to 3 parts: ca. 66 \% training, ca. 23 \% validation and ca 11 \% test. Then the reference dataset was rasterized to the same 1 m pixel grid as airborne dataset. Rasterization was performed after the train/validation/test split to ensure spatially correlated pixels of a single tree crown to be drawn only into one of the train/val/test parts.

\subsection{Neural network architecture}
We adopt a dual-stream multilayer perceptron (MLP) for pixel-wise land-cover classification from co-registered HSI and ALS data. 
\begin{figure}
\centering
\includegraphics[width=1.0\linewidth]{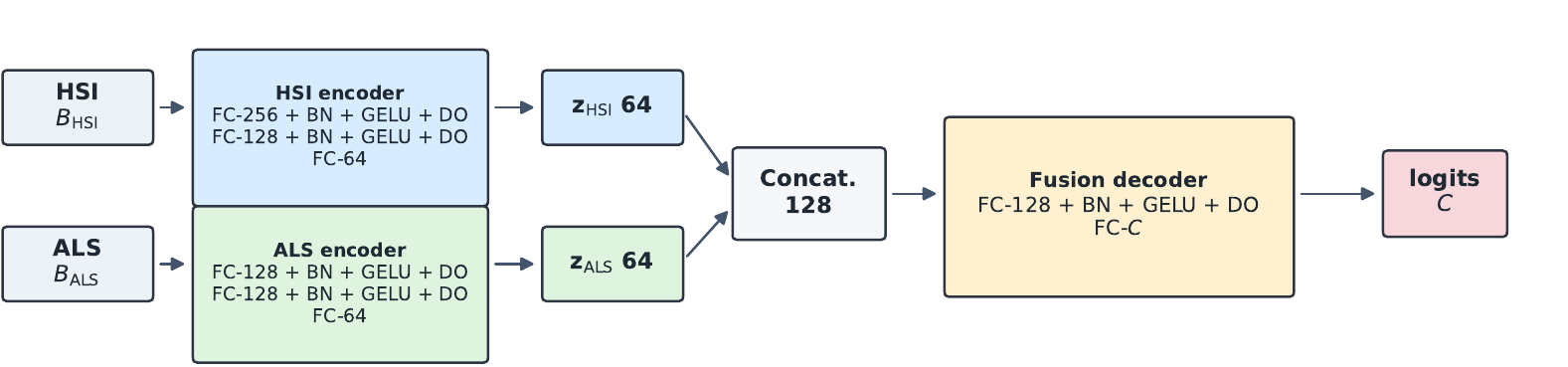}
        \caption{Dual-stream architecture used by DSNN. HSI and ALS vectors are encoded by separate MLPs with two normalized nonlinear blocks followed by a linear projection. The resulting modality-specific latent vectors are concatenated into a 128-dimensional representation and classified by a fusion decoder}
        \label{fig:neural_network}
    \end{figure} 
Each training sample consists of a spectral vector $\mathbf{x}^{\mathrm{HSI}} \in \mathbb{R}^{B_{\mathrm{HSI}}}$, an ALS feature vector $\mathbf{x}^{\mathrm{ALS}} \in \mathbb{R}^{B_{\mathrm{ALS}}}$, and a class label $y \in \{1,\dots,C\}$. The model comprises two modality-specific encoders followed by a fusion decoder and a final classifier, and is trained end-to-end on predefined train/validation/test splits.

This design follows a multi-view feature-fusion strategy: HSI and ALS are first encoded separately, allowing the network to learn modality-specific spectral and structural representations before combining them in a shared latent space for final classification.

The HSI and ALS encoders are implemented as separate MLPs,
\begin{equation}
f_{\mathrm{HSI}}: \mathbb{R}^{B_{\mathrm{HSI}}} \rightarrow \mathbb{R}^{D_{\mathrm{HSI}}},
\qquad
f_{\mathrm{ALS}}: \mathbb{R}^{B_{\mathrm{ALS}}} \rightarrow \mathbb{R}^{D_{\mathrm{ALS}}},
\end{equation}
each realized as a sequence of fully-connected blocks. Every hidden block consists of a linear layer, batch normalization, GELU (Gaussian Error Linear Unit \cite{hendrycks2016gaussian}) non-linearity, and dropout. The last encoder layer is a linear projection without normalization or dropout, yielding latent representations
\begin{equation}
\mathbf{z}_{\mathrm{HSI}} = f_{\mathrm{HSI}}(\mathbf{x}^{\mathrm{HSI}}),
\qquad
\mathbf{z}_{\mathrm{ALS}} = f_{\mathrm{ALS}}(\mathbf{x}^{\mathrm{ALS}}).
\end{equation}
These feature representations are concatenated into a joint feature vector
$\mathbf{z} = [\mathbf{z}_{\mathrm{HSI}}; \mathbf{z}_{\mathrm{ALS}}] \in \mathbb{R}^{D_{\mathrm{HSI}} + D_{\mathrm{ALS}}}$, which is passed to a fusion decoder MLP
\begin{equation}
g: \mathbb{R}^{D_{\mathrm{HSI}} + D_{\mathrm{ALS}}} \rightarrow \mathbb{R}^{C},
\end{equation}
constructed with the same block design (linear + batch normalization + GELU + dropout for hidden layers, followed by a final linear classification layer). The network outputs class logits
$\boldsymbol{\ell} = g(\mathbf{z}) \in \mathbb{R}^{C}$ and class probabilities
$p(c \mid \mathbf{x}^{\mathrm{HSI}}, \mathbf{x}^{\mathrm{ALS}})$ via a softmax.

\subsubsection{Hyperparameters}

The schema of the architecture is presented in Figure~\ref{fig:neural_network}. The HSI encoder consists of two hidden blocks followed by a 64-dimensional linear projection, with dimensions
$
B_{\mathrm{HSI}} \rightarrow 256 \rightarrow 128 \rightarrow 64,
$
and the ALS encoder has similar structure with dimensions
$
B_{\mathrm{ALS}} \rightarrow 128 \rightarrow 128 \rightarrow 64.
$
Thus, the concatenated latent representation has dimension
$D_{\mathrm{HSI}} + D_{\mathrm{ALS}} = 64 + 64 = 128$. The fusion decoder maps
$
128 \rightarrow 128 \rightarrow C,
$
where $C$ is the number of land-cover classes. 

We use Adam with learning rate $10^{-4}$ and weight decay $10^{-4}$, combined with a cosine-annealing schedule over the full training horizon. The model is optimised with an unweighted cross-entropy loss. During joint training, the two encoders and the decoder are updated simultaneously. In the results, this method is referred to as the Dual-Stream Neural Network (DSNN). 

For Wigierski dataset we used a dropout rate of $0.2$ applied after each hidden layer in both encoders and the decoder. Training was performed with a batch size(BS) of $512$ for $300$ epochs.  For Bory Tucholskie dataset we used dropout rate of $0.4$, and the training was performed for $500$ epochs with batch size 512.

\subsection{Cohabitation matrix}
\label{sec:cohabitation_matrix}
A symmetric cohabitation matrix $C \in [0,1]^{C \times C}$ where $C$ is the number of classes, is a prior describing compatibility of species in real forest habitats. It keeps $1$ on the diagonal (a species is always compatible with itself) and stores a real-valued association strength for every off-diagonal pair, reflecting how often those species are reported to occur together under comparable habitat conditions. 

In practice, it is estimated from habitat and vegetation reports (EUNIS databases \cite{chytry2020eunis}, literature \cite{pandey2023large} and expert knowledge), where repeated observations of species lists—often stratified by site descriptors like soil, moisture, elevation, stand age, and community type—allow for summarising co-occurrence as a graded affinity rather than a hard rule. The key idea is that a cohabitation matrix captures which mixtures are typical or rare in a specific environment. That makes it a useful tool for improving a semi-supervised scheme. Used in the pixel-growing algorithm, a cohabitation matrix can help correct the results toward more ecologically plausible neighbour expansions and reduce propagation of unlikely pseudo-labels.

We decided to use an automated approach to generating a cohabitation matrix.

\subsubsection{Cohabitation matrix generation}
\label{sec:cohab_matrix_generation}
We propose the LLM-driven approach to generating cohabitation information in the form of a matrix. This is aligned with \cite{dagdelen2024structured}, where authors show that large language models can reliably extract and structure relations from unstructured text when guided by constrained prompts and verification requirements.

Let $\mathcal{C}$ be a set of classes (tree species) and let $c = |\mathcal{C}|$. Let  $r>0$ denote the radius defining a neighbourhood within which trees may coexist. 

The cohabitation matrix $H_r \in [0,1]^{c \times c}$ is defined as
\begin{equation}
H_r[i,j] =
\begin{cases}
1, & i = j, \\
p_{ij}, & i \ne j,
\end{cases}
\end{equation}
where $p_{ij} \in [0,1]$ represents the likelihood that species $i$ and $j$ occur within distance $r$ of each other, where 1 denotes very
high likelihood and 0 denotes impossibility. The property that $H_r[i,i] = 1 \quad \text{for all } i \in \{1,\ldots,c\},$ reflecting that a species co-occurs with itself.

The cohabitation matrix was generated using a citation-constrained LLM workflow with web search enabled (ChatGPT 5.0 model). The LLM output was not treated as a scientific source; instead, the model was required to retrieve external habitat and co-occurrence sources, provide URLs or DOIs, and justify each matrix entry.  The prompt is provided in Section~\ref{sec:cohabitation_prompt}. The query can be customised with the following parameters:
\begin{itemize}
\item minimum number of sources per species: 2 -- number of required, documented sources for habitats in which each of the species is typically observed.
\item maximum number of sources per species: 4 -- number of required sources for observed co-existence for a given species.
\item distance in meters:  20 -- cohabitation distance
\item region: Central-Eastern Europe -- ecological context for the query and search for sources
\item additional environmental information: Wigry National Park, Poland. No large rivers or urban areas. Natural forests, agricultural areas and bogs.
\item species list: list of classes
\end{itemize}
To reduce the risk of hallucinated evidence, the prompt prioritised EUNIS databases and required verifiable URLs or DOI-indexed sources wherever possible. Visualization of the generated cohabitation matrix is presented in Figure~\ref{fig:cohabitation_matrix}. Note that, for source-identification purposes, matrix entries marked as $-1$ are replaced with $0$ when used by the algorithm.

Because the workflow relies on a non-deterministic LLM and on information retrieved from external sources, the resulting matrix should not be interpreted as a fixed or temporally invariant ecological ground truth. Rather, it represents an ecological hypothesis derived from the evidence available during a particular execution of the workflow and should be subject to expert verification.

\subsection{Cohabitation-aware pseudo-labels}
\label{sec:cohab_aware_pseudo}
We employ a two-pass classification framework. In the first pass, the initial classifier produces class probability estimates for all candidate tree-top pixels. In the second pass, the classifier's training set is augmented by candidate pixels (with pseudo-labels) selected from spatial neighbourhoods around the original labelled samples. Candidate pixels are selected based on the initial classifier output and a class co-occurrence prior that models species co-occurrence and spatial distance. In the results, this method is referred to as Dual-Stream Neural Network with Pseudo-labels (DSNN+P).

\subsubsection{Detecting treetops}
Candidate treetop pixels are identified with the Canopy Height Model (CHM) using a local maxima detection approach. The CHM is first preprocessed by clipping values to a valid height range (e.g., $[5, 40]$ m) and applying Gaussian smoothing ($\sigma=1.0$) to reduce noise. Local maxima are then extracted using a fixed window size ($5 \times 5$ pixels), with a minimum height threshold (e.g., $h_{min}=5$ m) to exclude ground vegetation. Flat plateaus are suppressed to ensure distinct peak localisation. The resulting binary map serves as the set of candidate locations $\mathcal{X}_{cand}$. 

Although CHM-based local maxima may also occur on elevated non-vegetation objects, such candidates are subsequently evaluated by the first-pass HSI--ALS classifier and retained only if their species-class confidence after cohabitation-aware fusion satisfies the pseudo-labelling criteria. In more heterogeneous or urban environments, this step could be further supported by additional filtering based on vegetation masks or geometric CHM/ALS criteria, such as local height variance, peak prominence, curvature, or point-cloud planarity.

\subsubsection{Selecting candidates}
Let $\mathcal{X}_{cand} = \{x_1, \dots, x_N\}$ be the set of candidate tree locations, and $\mathcal{X}_{train} = \{(x'_{1}, y'_{1}), \dots, (x'_{M}, y'_{M})\}$ be the set of labeled training samples (parents), where $y' \in \{1, \dots, K\}$ denotes the species class.

For each candidate $x_i$, the neural network classifier provides an initial vector of
species-class confidence scores:
\begin{equation}
\mathbf{P}_i = [p_{i,1}, \dots, p_{i,K}], \quad p_{i,k} \ge 0.
\end{equation}
These scores are the classifier probabilities for the $K$ pseudo-labelled
species classes. If the network contains additional background classes, these classes are omitted before the local fusion step.

We utilize a Cohabitation Matrix $\mathbf{C} \in \mathbb{R}^{K \times K}$, where $C_{uv}$ represents the frequency or likelihood of observing a neighbor of class $v$ given a central tree of class $u$.

To calibrate the influence of inter-species mixing versus intra-species clustering, we introduce an arbitrary scaling parameter $\delta$ (default $\delta=0.75$). We derive a scaled matrix $\mathbf{C}'$:
\begin{equation}
C'_{uv} =
\begin{cases}
C_{uv} & \text{if } u = v \\
\delta \cdot C_{uv} & \text{if } u \neq v
\end{cases}
\end{equation}
The parameter $\delta$ dampens off-diagonal compatibility values relative to the diagonal before row normalisation, reflecting that while inter-species cohabitation is possible, within-species clustering is often more likely.

This matrix is row-normalized to obtain the conditional prior distributions:
\begin{equation}
\mathbf{\pi}_u = [\pi_{u,1}, \dots, \pi_{u,K}], \quad \text{where } \pi_{u,v} = \frac{C'_{uv}}{\sum_{k} C'_{uk}}
\end{equation}

The algorithm iterates over every training sample (parent) and evaluates its influence on nearby candidates. For each candidate, the ``best'' parent (the one providing the strongest evidence) is selected to update the candidate's label.

\subsubsection{Local context fusion}
Let a single parent $j$ with location $x'_j$ and class $c_0 = y'_j$ be given. Let $x_i$ be a candidate within a maximum radius $R_{max}$, and the distance $d_{ij} = \|x_i - x'_j\|_2$.

We define a fused score vector $\mathbf{S}_{ij}$ that combines the classifier
scores $\mathbf{P}_i$ with the contextual prior $\mathbf{\pi}_{c_0}$ using a
distance-dependent mixing coefficient. The coefficient is
$\lambda_{ij}=w(d_{ij})$, unless a fixed mixing coefficient is used.

The distance function $w(d)$ models the decay of contextual influence:
\begin{equation}
w(d) =
\begin{cases}
1 & \text{if } d \le R_{min}, \\
\epsilon + (1-\epsilon)\sqrt{1 - \left(\frac{d-R_{min}}{R_{max}-R_{min}}\right)^2} & \text{if } R_{min} < d < R_{max}, \\
\epsilon & \text{if } d \ge R_{max}.
\end{cases}
\end{equation}
Thus candidates close to a labelled parent are influenced more strongly by the
cohabitation prior, while candidates near the edge of the search radius are
influenced mostly by the first-pass classifier.

For class $k$, the mixed score is
\begin{equation}
S_{ij,k} =
(1-\lambda_{ij}) p_{i,k}
+ \lambda_{ij} \pi_{c_0,k}.
\end{equation}
The score vector is then normalized:
\begin{equation}
\tilde{S}_{ij,k} =
\frac{S_{ij,k}}{\sum_{\ell=1}^{K} S_{ij,\ell}}.
\end{equation}

\subsubsection{Candidate selection}
After computing the normalized mixed scores for all labelled parents in the
neighbourhood of candidate $i$, the selected parent is the one that yields the
largest single-class confidence:
\begin{equation}
j^* = \arg\max_{j \in \mathcal{N}(i)}
\left( \max_{k} \tilde{S}_{ij,k} \right).
\end{equation}
The pseudo-label and its confidence are then
\begin{equation}
\hat{y}_i = \arg\max_k \tilde{S}_{ij^*,k},
\quad
\hat{p}_i = \max_k \tilde{S}_{ij^*,k}.
\end{equation}
 
\subsubsection{Augmented dataset construction}
\begin{figure*}
    \centering
    \begin{subfigure}{0.32\linewidth}
        \centering
        \includegraphics[width=\linewidth]{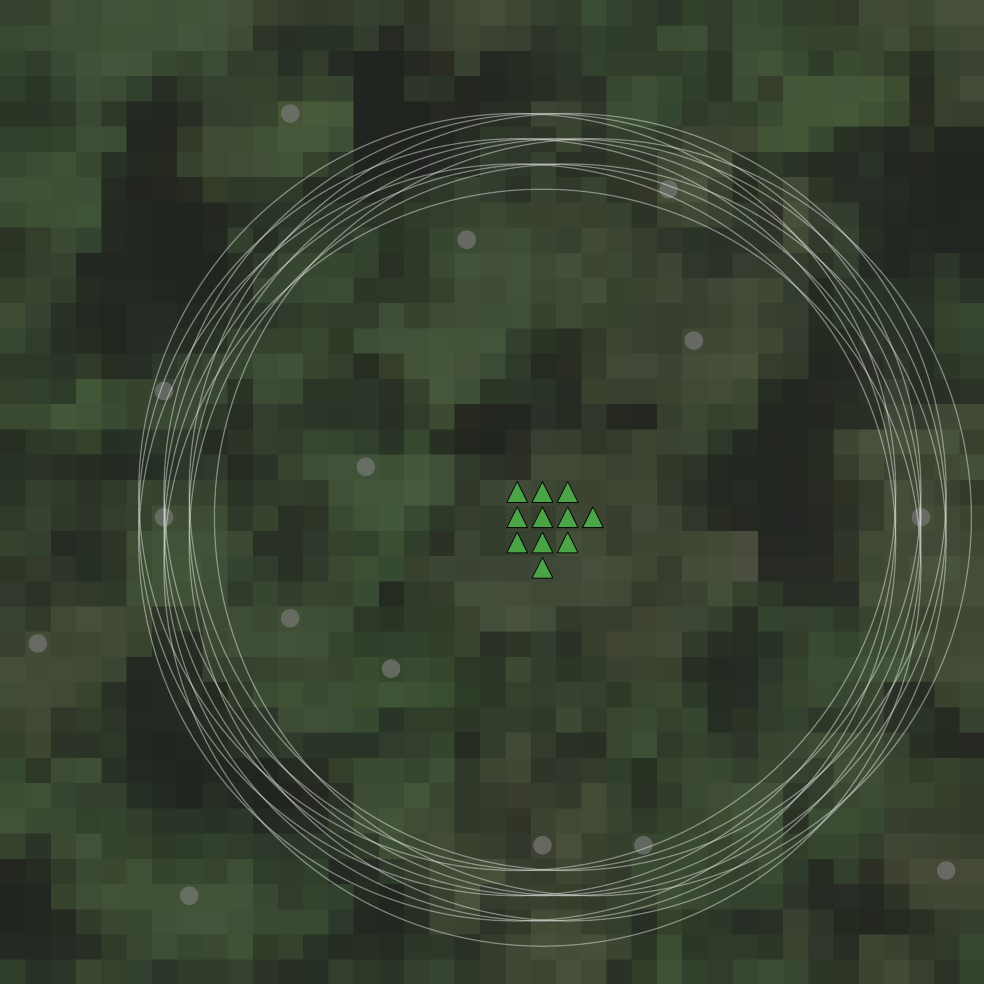}
        \caption{}
    \end{subfigure}
    \begin{subfigure}{0.32\linewidth}
        \centering
        \includegraphics[width=\linewidth]{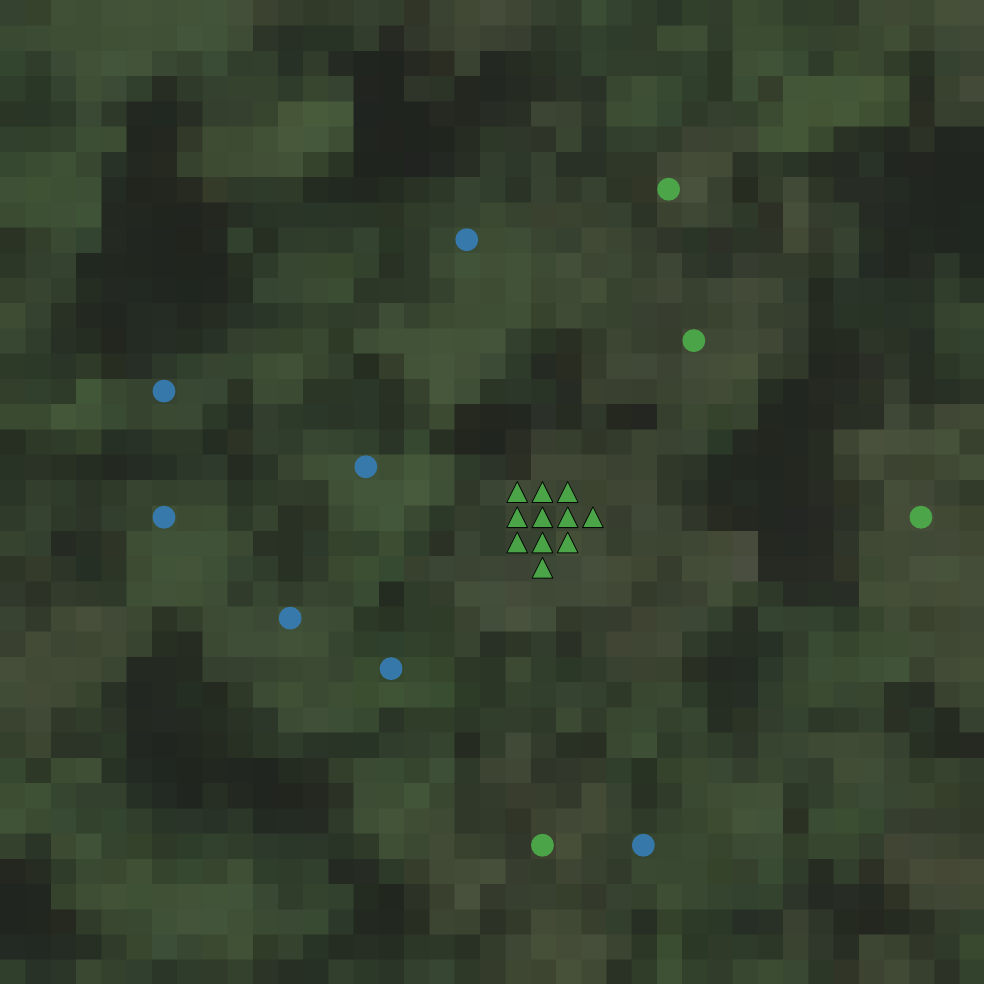}
        \caption{}
    \end{subfigure}
    \begin{subfigure}{0.32\linewidth}
        \centering
        \includegraphics[width=\linewidth]{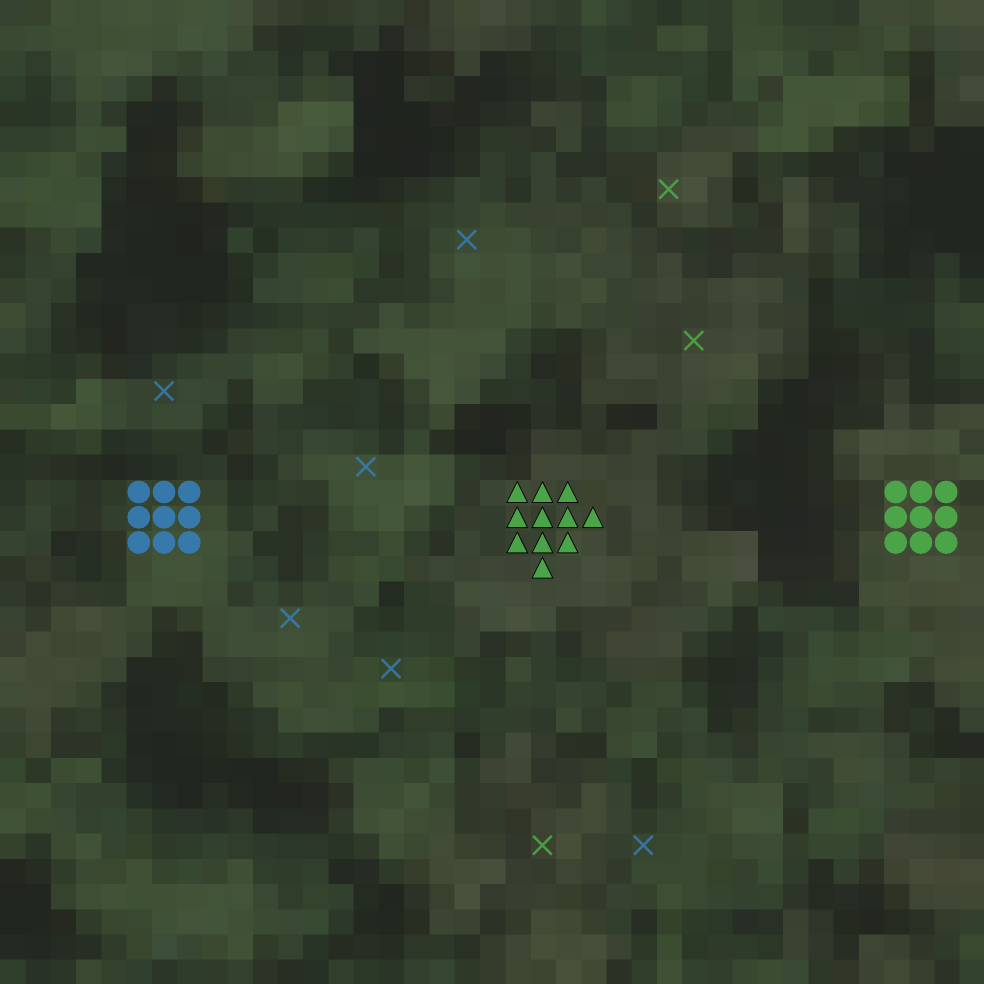}
        \caption{}
    \end{subfigure}
    \caption{Visualization of the pseudo-label algorithm: (a) treetops detected around training (parent) pixels; (b) selected candidates with assigned classes; (c) candidates denoted with a cross are filtered out with probability threshold, remaining candidates are subject to spatial expansion. Green denotes the class \textit{Betula }spp., and blue denotes the class \textit{Alnus glutinosa}.}
    \label{fig:cohabitation_examples}
\end{figure*}

The refined pseudo-labels are used to generate an augmented training dataset. 
To do this, we retain only those refined candidates whose assigned class probability exceeds the high-confidence threshold $\hat{p}_i \ge \tau$. In the  experiment with dataset Wigierski, we used $\tau=0.99$; for the dataset Bory Tucholskie we
used $\tau=0.85$. Candidates located at coordinates already present in the training set are discarded. Visualization of the algorithm is presented in Figure~\ref{fig:cohabitation_examples}.

We note that this pseudo-label confidence threshold $\tau$ is dataset dependent. This indicates sensitivity to this parameter, and a more systematic analysis of its influence constitutes an important direction for future evaluation.

We also increase the size of the training set by performing neighbourhood expansion for every candidate, returning its $ n$-neighbourhood with the same label as the central candidate (e.g., $n=1$ yields a $3 \times 3$ block). Coordinate overlaps are resolved by retaining centre pixels
before expanded neighbours and then keeping the first remaining occurrence.

\subsubsection{Error propagation problem}
The proposed pseudo-labelling procedure was designed to reduce error accumulation and confirmation bias, which are common risks in semi-supervised learning. Pseudo-labels are assigned only to ALS-derived treetop candidates, where mixed spectra and crown-boundary effects are expected to be weaker than at canopy edges. Candidate acceptance depends not only on classifier confidence, but also on spatial proximity to labelled trees and the ecological cohabitation prior. In addition, pseudo-labels are retained only when they exceed a high-confidence threshold. The procedure is applied only once and is not iterative: newly added pseudo-labelled pixels are not used to generate further pseudo-labels, and all candidate selection is based on the original, field-verified training data. These safeguards do not eliminate pseudo-label noise, but they limit its propagation in the evaluated datasets.

\subsection{Reference methods}
\label{sec:reference_methods}
Reference methods follow a classical pipeline of feature extraction and classification. Three feature sets were prepared: the first, the original hyperspectral reflectance cube (HSI). The second feature set was created using two feature extraction methods applied to the HSI mosaic:  30 most informative bands of the minimum Noise Transformation (MNF)~\cite {Jarocinska2024}; spectral indices (IND) derived from the HSI mosaic using ENVI SpectraIndices~\cite{ENVIApiGuide}. These features were combined into a second dataset referred to as MNF-IND. The complete list of calculated indices is included in Table~\ref{tab:hs_ind} in the Appendix. The third feature set included features derived from ALS point clouds using the lidR library~\cite{lidRbook}, combined into the MNF-IND-ALS dataset. ALS features were calculated as statistical metrics from the point cloud in the 1 m grid of HSI imagery. The complete list of calculated ALS metrics is presented in Table~\ref{tab:lidar_metrics}. Those three datasets represent classical approaches to creating airborne remote sensing datasets for vegetation classification and correspond to established strategies for HSI classification:  (1) HSI consists of the original data without additional feature engineering and is used to ensure a fair comparison between reference methods and the proposed approach on identical inputs; (2) MNF-IND represents the improvement resulting from state-of-the-art feature extraction; (3) MNF-IND-ALS extends this by integrating an additional data modality. 

The reference classification method was the pixel-based CatBoost~\cite{Prokhorenkova2018} algorithm. This algorithm was successfully used in previous works on tree species classification~\cite{niedzielko_airborne_2024}. The model training was performed on all training pixels extracted from rasterized reference dataset. Three models were trained, using 3 datasets: HSI, MNF-IND, MNF-IND-ALS. Internal model validation was performed on pixels extracted from validation dataset. Accuracy assessment was done on the test dataset. Experiments were repeated five times, and results were averaged. Scenarios tested in experiments are presented in Table~\ref{tab:scenarios}. As our main accuracy measure, we used the F1 score computed for each class (per-class F1) and their arithmetic mean (macro-averaged F1 score) as the aggregated metric. 
The maps obtained under the best scenarios were compared using the Jaccard index, defined as the ratio of the area of intersection between regions assigned to the same classes and the total area covered by their union.

\begin{table*}[ht]
    \centering
    \begin{tabular}{lll}
        \toprule
        Scenario acronym& Classifier & Description \\
        \midrule
        HSI &
        CatBoost &
        Reflectance\\

        MNF-IND &
        CatBoost &
        MNF bands, spectral indices from reflectance\\

        MNF-IND-ALS &
        CatBoost &
        MNF bands, spectral indices from reflectance, metrics from the ALS point cloud\\

        DSNN &
        Dual-Stream Neural Network &
        Reflectance, metrics derived from the ALS point cloud. \\

        DSNN+P &
        Dual-Stream Neural Network &
        Reflectance, metrics derived from the ALS point cloud, pseudo-labels. \\
        \bottomrule
    \end{tabular}
    \caption{Classification scenarios and algorithms used in experiments.}
    \label{tab:scenarios}
\end{table*}

\section{Results}
\label{sec:results}
In our experiments, the proposed Dual-Stream Neural Network with Pseudo-labels (DSNN+P) algorithm reached $79.38\pm0.8$ macro F1, $90.09\pm0.3$ accuracy and $81.87\pm0.04$ balanced accuracy (mean recall). 
Classification results of the proposed approach (DSNN and DSNN+P) and reference methods using vegetation indices, feature extraction and classification algorithms are presented in Table~\ref{tab:class_metrics} (for scenarios see Table~\ref{tab:scenarios}) . DSNN+P provides a large increase in average accuracy compared to the reference methods, and the proposed semi-supervised step yields the best average result and is often superior for in-class classification. The observed gains in macro-F1 are 36.84, 8.35, and 6.01 percentage points, respectively, compared with HSI, MNF-IND, and MNF-IND-ALS. Pseudo-labelling improves macro-F1 by 1.96 percentage points compared with DSNN.

Our choice of reference methods (see Section~\ref{sec:reference_methods}) allows us to quantify the gains delivered by DSNN/DSNN+P against typical HSI and HSI+ALS processing pipelines. The large improvement of MNF-IND over raw HSI ($+28.49$ macro F1) is consistent with the well-known difficulty of learning from high-dimensional hyperspectral inputs without dedicated feature extraction. In contrast, adding ALS to the classical pipeline yields only a modest additional gain (MNF-IND-ALS vs. MNF-IND $+2.34$ macro F1), suggesting limited benefit from simple feature level fusion. DSNN/DSNN+P achieves a substantially larger improvement, indicating that it learns a more effective joint representation from both HSI and ALS.

\begin{figure*}
    \centering
    \includegraphics[width=1.0\linewidth]{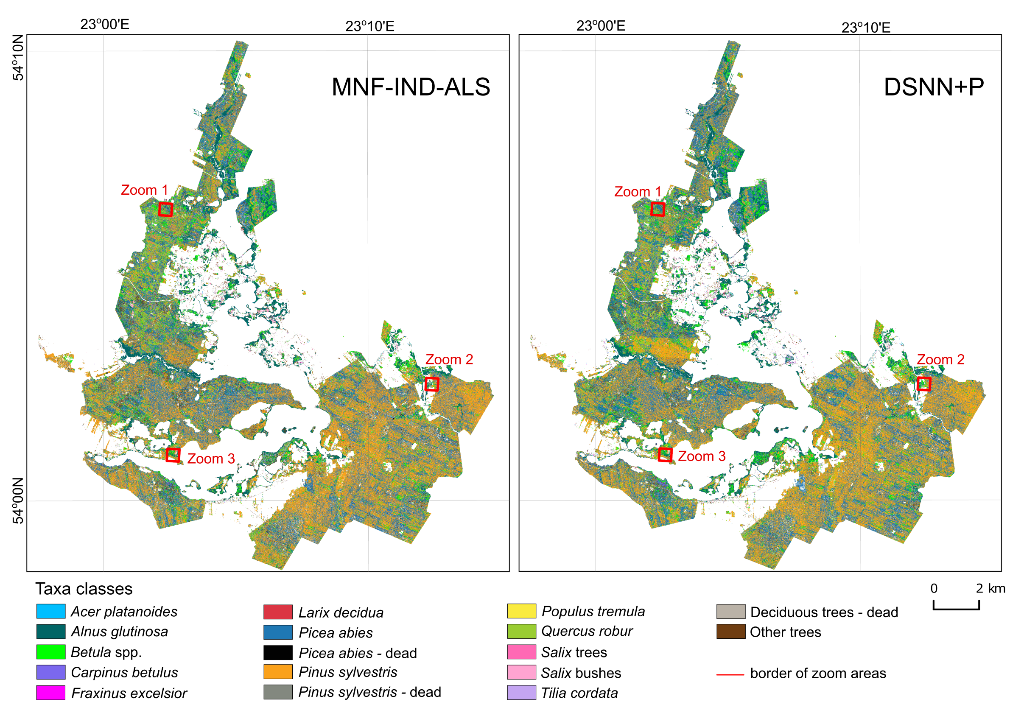}
    \caption{Tree taxa maps for the entire study area generated using two of the compared methods (MNF-IND-ALS and DSNN+P). The areas shown in detail in Figure~\ref{fig:yroi_contours} are indicated on the map by three squares.}
    \label{fig:xtwoareas}
\end{figure*}

We compared the class-level results of the best reference method (MNF-IND-ALS) with those of the proposed DSNN+P architecture. Comparisons of class-specific spatial patterns across the entire study area show high agreement between the maps generated by the two approaches, consistent with the numerical results (Figure~\ref{fig:xtwoareas}) and confirmed by a Jaccard index of 0.939. 
Class proportions are also comparable (Figure~\ref{fig:zperc_contrib}).
\begin{figure*}
        \centering
        \includegraphics[width=1.0\linewidth]{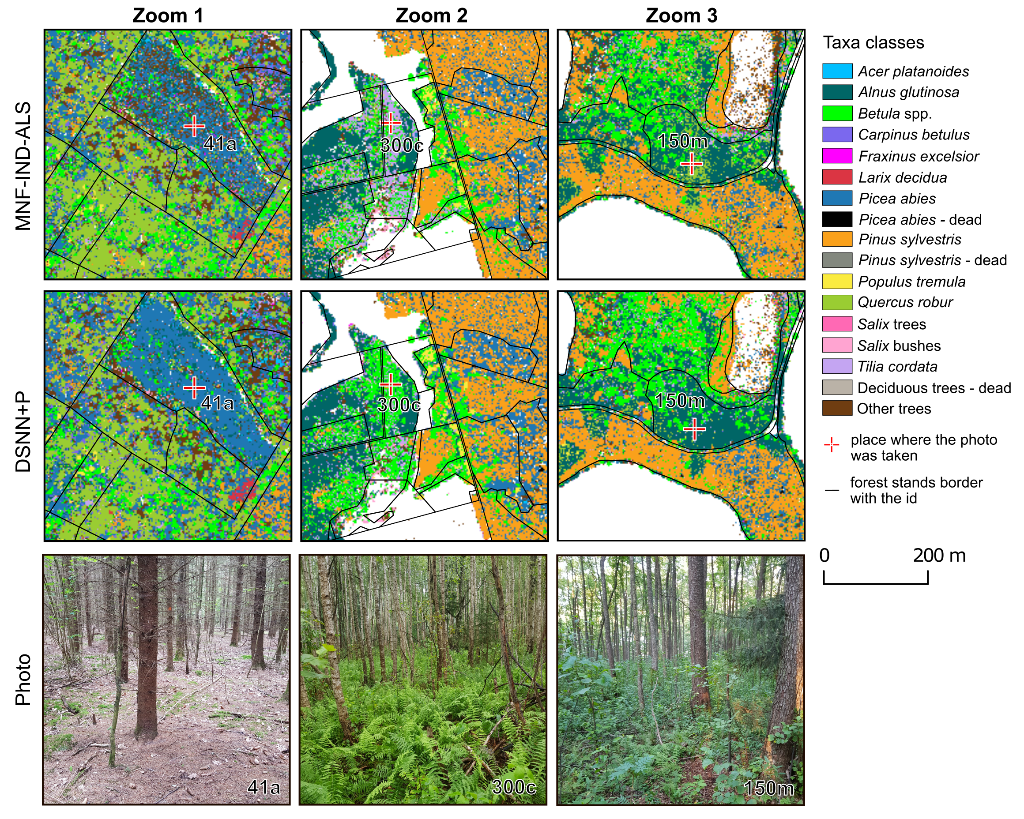}
        \caption{Map of tree taxa generated with the two compared methods (MNF-IND-ALS and DSNN+P) for three selected zoom-in areas. The zooms were chosen in locations showing the largest discrepancies between the maps. For each area, a field photograph was taken; the photo location is indicated by a red cross.}
        \label{fig:yroi_contours}
    \end{figure*}     
    \begin{figure}
        \centering
        \includegraphics[width=1.0\linewidth]{fperc_contrib_small.png}
        \caption{Comparison of the area covered by trees for different taxon classes, calculated from maps generated using the DSNN+P and MNF-IND-ALS scenarios. Percentages are given as labels for classes that cover > 1\% of the area.}
        \label{fig:zperc_contrib}
    \end{figure} 
The largest difference in percentage of occurrence is for the \textit{Alnus glutinosa} class, at 2.11 percentage points. Discrepancies are observed locally in small areas (examples on Figure~\ref{fig:yroi_contours}). In the first column(Zoom 1, Inventory precinct: Leszczewk, forest division and separation: 41a), the forest stand description~\cite{fdb2026} indicates a pure 38-year-old Norway spruce (\textit{Picea abies}) stand, also confirmed by the field photo; thus, \textit{Picea abies }should be almost the only class detected. DSNN+P better matches this condition, whereas MNF-IND-ALS assigns numerous trees to the \textit{Other trees} class, which is present  in the area, but occurs in much smaller proportions. In the second column (Zoom 2, inventory precinct: Wysoki Most, forest division and separation: 300c), DSNN+P again aligns more closely with the field situation by mapping a pure birch stand (\textit{Betula} spp.) at the photo location; in contrast, MNF-IND-ALS identifies small-leaved linden \textit{Tilia cordata} in substantial quantities despite its absence on site. A similar pattern is observed in the third column (Zoom 3, inventory precinct: Sobolewo, forest division and separation: 150m), where oak (\textit{Quercus robur}) does not occur, and black alder (\textit{Alnus glutinosa}) and birch (\textit{Betula }spp.) dominate; DSNN+P provides a markedly more accurate map. Beyond improved aggregate metrics, visual inspection against field photographs in selected areas confirms more accurate labelling at the level of individual crowns and small tree groves.

\begin{figure}
    \centering
    \includegraphics[width=1.0\linewidth]{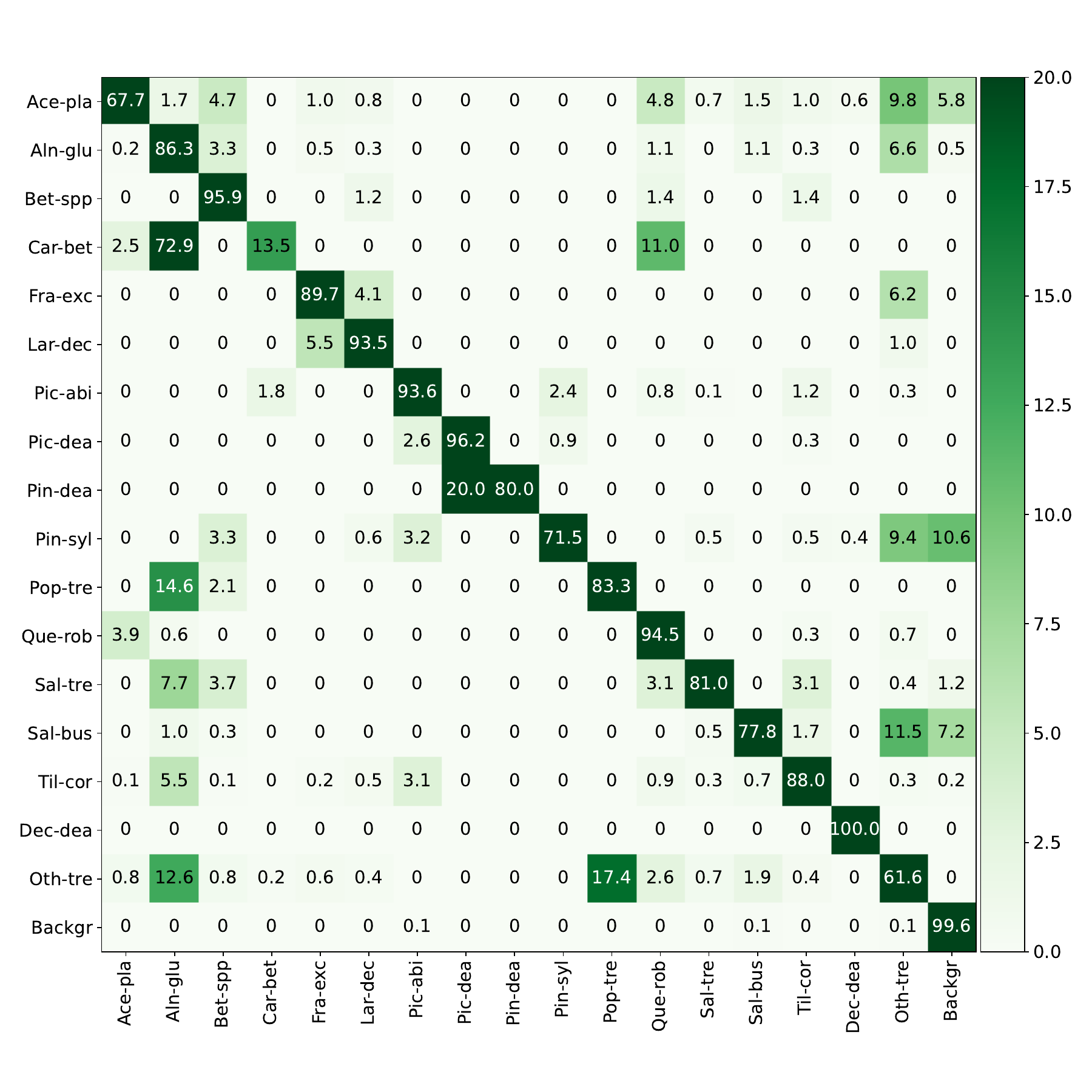}
    \caption{Confusion matrix for DSNN+P method averaged over experiments}
    \label{fig:avg_cm}
\end{figure}

Regarding individual tree species, the confusion matrix of the DSNN+P method, averaged over all experiments and row-normalized is presented in Figure~\ref{fig:avg_cm}. 
Overall, 12 classes achieve recall above 80\% (10, not counting the last three classes that aggregate diverse targets). The most confusing tree taxa classes (measured by class-wise error rate: $1-$recall above 25\%) are: \textit{Carpinus betulus} (86.46\% error), \textit{Acer platanoides} (32.28\%) and \textit{Pinus sylvestris} (28.54\%) in addition to the \textit{Other trees} class (38.44\%). 

Some classes are frequently predicted for other taxa, i.e. they `attract' false positive errors. We characterise this tendency using the aggregated incoming error AGE, defined as the off-diagonal column sum in the confusion matrix, and the number of classes NC
with incoming confusion above a threshold $d=1$\%.
Ordering classes by AGE and NC, the most prominent false-positive attractors are: \textit{Alnus glutinosa} ($AGE=116.7$\%, $NC=7$), Other trees ($AGE=46.2$\%, $NC=5$),  Background ($AGE=25.3$\%, $NC=4$) and \textit{Quercus robur} ($AGE=25.6$\%, $NC=6$).

\begin{table*}
		\centering
		\begin{tabular}{llrrrrrr}
			\toprule
			Tree taxa class & Class code & HSI & MNF-IND & MNF-IND-ALS & DSNN & DSNN+P & class-counts \\
			\midrule
			\textit{Acer platanoides }& Ace-pla & 37.69 & 63.63 & 70.80 & 73.76 &\textbf{75.25}& 145 \\
			\textit{Alnus glutinosa} & Aln-glu & 40.16 & \textbf{60.14} & 55.94 & 53.16 &55.24& 133 \\
			\textit{Betula} spp. & Bet-spp & 43.60 & 77.16 & 65.92 & 83.57 &\textbf{84.19}& 83 \\
			\textit{Carpinus betulus} & Car-bet & 0.80 & 17.28 & 10.47 & \textbf{26.10} &22.73& 96 \\
			\textit{Fraxinus excelsior} & Fra-exc & 15.00 & 44.06 & 74.19 & 80.67 &\textbf{83.65}& 29 \\
			\textit{Larix decidua} & Lar-dec & 49.58 & 63.03 & 87.21 & \textbf{87.40}&87.06& 40 \\
			\textit{Picea abies} & Pic-abi & 69.92 & 91.47 & 91.06 & 91.53 &\textbf{92.22}& 238 \\
			\textit{Picea abies} - dead & Pic-dea & 89.50 & \textbf{97.78} & 97.52 & 96.21 &97.35& 69 \\
			\textit{Pinus sylvestris} & Pin-syl & 60.17 & 81.23 & 76.53 & 80.81 &\textbf{88.89}& 157 \\
            \textit{Pinus sylvestris} - dead & Pin-dea & 0.00 & \textbf{97.78} & 86.11&81.13&80.80& 5 \\
			\textit{Populus tremula} & Pop-tre & 27.99 & 53.33 & 60.54 & \textbf{68.38} &67.13& 96 \\
			\textit{Quercus robur} & Que-rob & 67.01 & 86.19 & 86.85 & 90.14 &\textbf{91.30}& 305 \\
			\textit{Salix} trees & Sal-tre & 41.27 & 43.97 & 56.53 & 85.77&\textbf{86.36}& 104 \\
			\textit{Salix} bushes & Sal-bus & 14.41 & 70.91 & \textbf{86.90} & 81.11 &80.95& 117 \\
			\textit{Tilia cordata} & Til-cor & 70.94 & 88.34 & 89.26 & 90.55&\textbf{91.96}& 438 \\
			Deciduous trees - dead & Dec-dea & 10.00 & \textbf{86.67} & 60.00 & 68.67 &75.43& 2 \\
			Other trees & Oth-tre & 34.02 & 57.73 & 65.54 & 68.35      &\textbf{68.98}& 359 \\
			Background  & Backgr & 93.70 & 97.81 & \textbf{99.22} & 99.05   &99.11& 2637 \\
			$\bar{x}$ & & 42.54 &71.03  &73.37  &77.42&\textbf{79.38}&	\\
			\bottomrule
		\end{tabular}
        \caption{Results comparison for tested methods. Values are per-class F1 scores (reported as percentages, $\times 100$) and macro-averaged F1 ($\bar{x}$, bottom row). For the proposed Dual-Stream Neural Network (DSNN) and DSNN with Pseudo-labels (DSNN+P), results are averaged over five runs.}
        \label{tab:class_metrics}
	\end{table*}

\begin{figure}
    \centering
    \begin{subfigure}{0.49\linewidth}
        \centering
        \includegraphics[width=\linewidth]{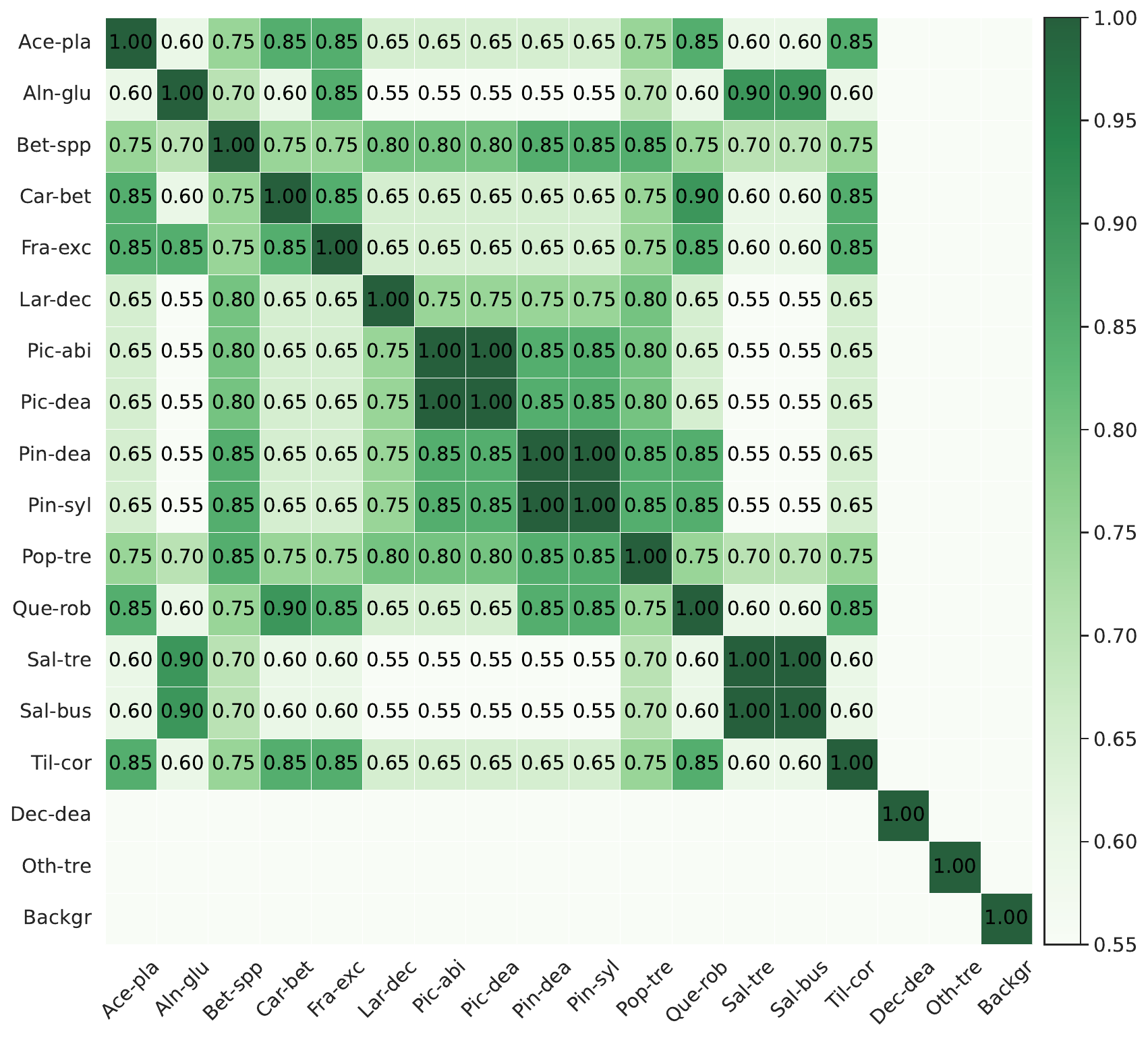}
        \caption{Cohabitation matrix}
        \label{fig:classes}
        \label{fig:cohabitation_matrix_a}
    \end{subfigure}
    \hfill
    \begin{subfigure}{0.49\linewidth}
        \centering
        \includegraphics[width=\linewidth]{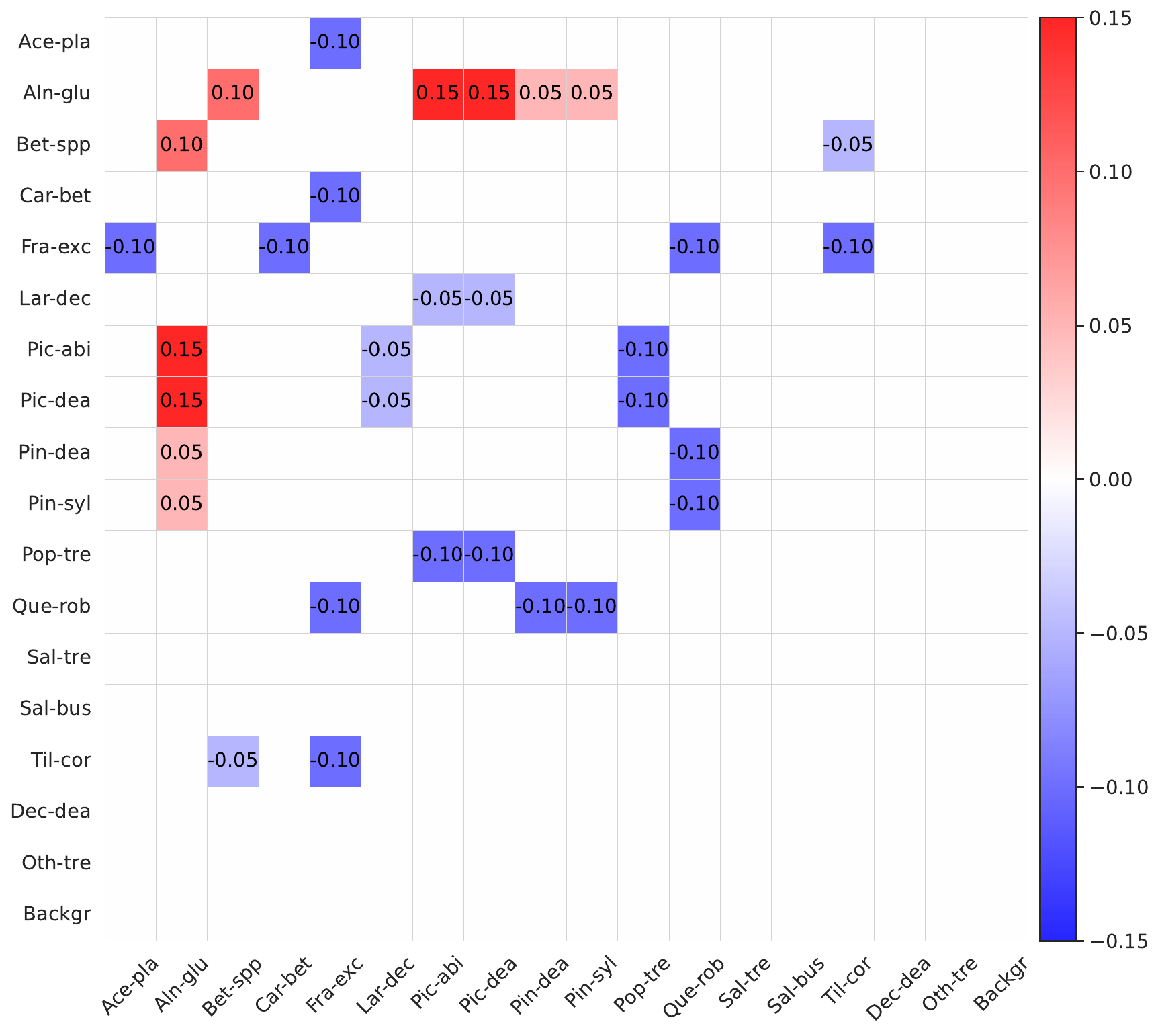}
        \caption{Expert updates}
        \label{fig:sceneF}
        \label{fig:cohabitation_matrix_b}
    \end{subfigure}
    \caption{LLM-based cohabitation matrix (a) and differences proposed by experts (b). Values are percentages, cells without value should be treated as zero. Wigierski dataset.}
    \label{fig:cohabitation_matrix}
\end{figure}

\subsection{Cohabitation matrix evaluation} 
\label{sec:Cohabitation_evaluation}
Generating the cohabitation matrix required the LLM to analyse each unordered pair of species separately.  We decided that the key factor in assessing realistic cohabitation estimates is the existence of supporting sources (EUNIS and environmental databases, as well as relevant scientific papers) for both species in each pair inhabiting the same or very similar habitats. For each pair, the model was directly instructed to provide natural habitats with believable sources of habitat and cohabitation. The model was instructed to decrease the cohabitation coefficient if supporting sources could not be found (see part 4 , Scoring guidance'' of the prompt in Section \ref{sec:cohabitation_prompt}).
We specifically required sources where both species in a pair appear together -- if such could be found -- which would be a stronger argument for trees growing in relatively close proximity ($r=20$) required for the cohabitation matrix. 

The cohabitation matrix created by the LLM-based system is presented in Figure~\ref{fig:cohabitation_matrix_a}. This matrix was verified by domain experts, familiar with forest ecology in north-eastern Poland and the local conditions of Wigry National Park. They evaluated the matrix in the context of the habitats in the study area and at the spatial scale corresponding to the adopted neighborhood range (20 meters). The adjustments proposed by them are presented in Figure~\ref{fig:cohabitation_matrix_b}. Changes in the cohabitation matrix affected 13 of the 18 taxa. Expert changes in the matrix ranged from -0.10 to +0.15. The dominant direction of change was a decrease in probability by -0.10. The largest change, +0.15, concerned the probability of occurrence in the neighbourhood of the species \textit{Picea abies} (code Pic\_abi and Pic\_dea) and \textit{Alnus glutinosa (Aln\_glu)}. For these taxa, the original matrix indicated a probability of 0.55, which was clearly insufficient given the frequent occurrence of Norway spruce as an admixture species with permanent coverage in stands dominated by black alder in north-eastern Poland~\cite{matuszkiewicz2012}

\section{Discussion}
\label{sec:discussion}
Our results are comparable to those reported in previous studies. \cite{weinstein2024individual} employs a hierarchical, multi-temporal deep learning model with spectral attention mechanisms for landscape-level HSI and ALS-based detection and classification of over 100 million individual trees across 24 sites from the National
Ecological Observatory Network (NEON) dataset, achieving the average macro-accuracy of 75.8\%. Assuming that macro-accuracy is comparable to balanced accuracy, this is (6.07) percentage points lower than the balanced accuracy obtained by DSNN+P on the Wigierski dataset ($81.87 \pm 0.04$\%). However, because the datasets differ, this comparison can only be interpreted as an indirect reference point.
Similarly, \cite{quan2023tree} proposed a workflow for HSI- and ALS-based tree species classification, using recursive feature selection and extraction coupled with a Random Forest model. On the dataset of secondary forest in Northern China, this approach achieved an  average macro-F1 of 71.5\%, again below our reported $(79.38 \pm 0.8) \%$. 
In addition, \cite{graves_data_2023} reports low accuracy for rare or less numerous species, a macro-F1 value of approximately $0.32$.
Therefore, while our study uses a different dataset, these comparisons suggest that our approach represents a promising method for tree species classification.  

The relatively modest capacity of DSNN is a deliberate choice given the limited number of independently labelled trees. Although deeper or more highly parameterised architectures can provide greater representational capacity, under sparse supervision this advantage must be balanced against increased susceptibility to overfitting and unstable generalisation. This trade-off is particularly relevant in hyperspectral tree-species classification, where previous studies have reported substantial degradation of CNN-based classifiers when the number of training samples is restricted, while simpler models remain comparatively robust~\cite{tong2022spectral}. We therefore use a compact dual-stream MLP that retains sufficient capacity to learn separate HSI and ALS representations and their joint fusion, while limiting model complexity relative to larger deep architectures.
The proposed methodology does not employ label regularisation (e.g. majority voting among neighbouring pixels). Although spatial smoothing can improve results in some areas (Figure~\ref{fig:yroi_contours}), its effect is not uniformly beneficial and may blur fine-scale boundaries; therefore, we report unregularised predictions to preserve pixel-level detail.

Indirect comparisons of accuracy across different datasets are inherently uncertain due to differences in site conditions, label quality, class composition, and acquisition settings. We therefore compare species-mapping methods under a common experimental protocol, in which DSNN+P yields higher accuracy and produces maps that better align with expert visual assessments.

\subsection{Methodological motivation and applicability}
The proposed methodology addresses a practical challenge frequently encountered in ecological remote sensing: obtaining reliable tree-level annotations for multi-modal datasets. Notable approaches to such annotation challenges include synthetic data generation and automatic annotation from external sources, such as Harvester Production Reports (HPRs). Synthetic data have been shown to improve classification performance in hyperspectral imagery~\cite{yang2026fedda} and LiDAR-based forest applications~\cite{jarahizadeh2026smote}; however, these approaches generally operate on a single sensing modality and focus on augmenting existing labelled datasets rather than acquiring new annotations. Alternatively, HPRs can provide automatic tree-level labels for remote sensing data~\cite{de2025boreal}, but such information is available only in forests where mechanised harvesting operations are conducted.

The practical value of our framework is therefore greatest in environments where annotation resources are limited, harvesting-derived labels are unavailable, and ecological knowledge can be used as an additional source of information. These conditions are characteristic of protected forest ecosystems such as Wigierski National Park, where mechanised harvesting is unavailable and extensive field campaigns are costly, logistically constrained, and limited by conservation requirements. In such settings, species co-occurrence patterns documented in the ecological literature and in habitat databases such as EUNIS~\cite{chytry2020eunis} provide an alternative source of prior knowledge to guide pseudo-label assignment. Unlike purely confidence-based pseudo-labelling strategies~\cite{bryson2024domain}, our approach explicitly constrains candidate labels using ecologically plausible species associations. Importantly, the proposed LLM-driven extraction of cohabitation priors is scalable in the sense that it can be initiated from a species list, a geographic or ecological region, and appropriately constrained literature and habitat sources. However, its use still depends on sufficient ecological evidence and should be accompanied by expert verification where possible. Consequently, the framework is not limited to the study area considered here and can, in principle, be transferred to any forest ecosystem for which sufficient ecological knowledge is available. This makes the methodology particularly attractive for semi-supervised tree species mapping in data-scarce, ecologically complex, and multi-sensor forest inventories.

\subsection{Cohabitation matrix}
None of the expert corrections to LLM-generated matrix was lower than $-0.10$ and higher than $0.15$.
From a botanical point of view, the generated matrix can therefore be considered approximately correct and consistent with botanical domain knowledge of the co-occurrence of species in the forests of north-eastern Poland~\cite{matuszkiewicz2012}. 

As mentioned in Section \ref{sec:Cohabitation_evaluation}, the expert evaluation shows that the LLM-generated matrix achieves its main goal: capturing the region's ecological structure. Consistent with the prompt, high values are generally assigned to species with comparable habitat requirements, while low values are associated with clear habitat mismatches, and the experts confirmed the general results as acceptable. 

Most corrections targeted borderline cases where “regional co-occurrence” can be misleading without finer habitat stratification (e.g., wet vs. mesic sites, peatland edges, or successional transitions), as well as classes defined at genus or functional levels (e.g., \textit{Betula} spp., \textit{Salix} trees vs. bushes), where a single label aggregates multiple ecological niches and thus blurs pairwise relationships. A separate group concerns dead-tree classes (e.g., \textit{Picea} dead, \textit{Pinus} dead), for which spatial adjacency reflects stand structure and disturbance processes rather than genuine habitat preference; consequently, such entries should be interpreted as weaker priors. 

Another approach to replace LLM  queries is to infer the cohabitation matrix directly from the training data. This, however, would be unreliable for the majority of similar photogrammetric images of tree cover datasets due to the sparsity of the training data. Individual labelled samples (trees) are sparsely distributed, and there are only a limited number of scenarios in which two or more samples are located within a small radius (20 meters in this work). This sparse distribution would make the estimated likelihoods of close neighbourhoods prone to large errors.

Importantly, in our framework, the matrix is not treated as a calibrated ecological probability model (and certainly not a full replacement of experts) but as a soft compatibility prior that modulates pseudo-label selection on the canopy graph. Therefore, the most critical property is the preservation of relative plausibility (ranking of more vs. less likely neighbours) rather than exact numerical calibration. 
Its main advantage is providing a scalable and traceable way to construct an initial, citation-grounded ecological prior when dense field-based co-occurrence data are unavailable.
The quality of this output is controlled through a constrained prompt requiring species lists, regional context, neighbourhood radius, verifiable ecological sources, score justifications, and, where possible, expert verification.

An additional limitation concerns the reproducibility and temporal stability of LLM-derived priors. Repeated execution of the workflow may produce differing results due to stochastic model behaviour, changes in the underlying LLM and retrieval mechanisms, and changes in the available scientific literature as new evidence is published. The resulting matrix should therefore be regarded as an ecological hypothesis associated with a documented execution of the workflow rather than as a fixed ground-truth representation. Retaining the retrieved sources, pairwise rationales, and generated scores makes this process traceable and allows its outcome to be independently assessed by domain experts. Future work should investigate open-source LLMs with fixed model versions (especially open source) and deterministic sampling settings, which would allow stricter replication of the knowledge-extraction procedure across environments and time points.

\subsection{Verification on an independent dataset}
\begin{figure}
        \centering
        \includegraphics[width=1.0\linewidth]{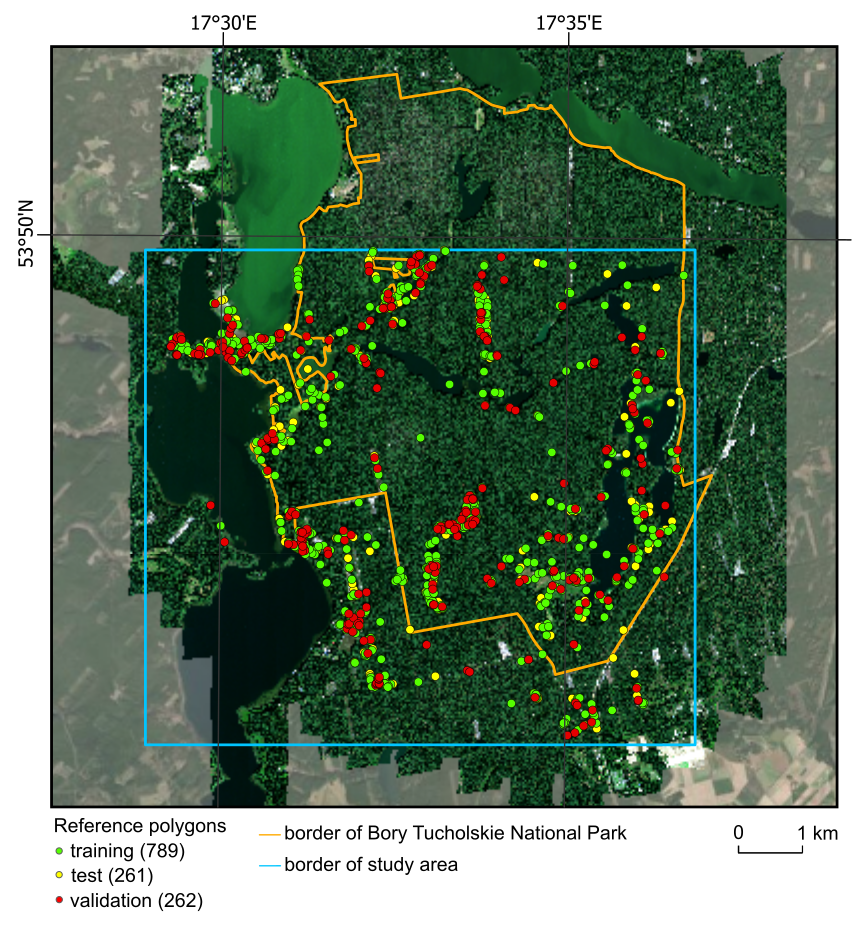}
        \caption{Visualisation of the Bory Tucholskie dataset with sampled trees divided into training, validation, and tests subsets}
        \label{fig:study_area_bt}
    \end{figure}

To verify our results, we employed a secondary dataset. It contains data acquired over Bory Tucholskie National Park, with the same sensor configuration as the primary dataset (see Section~2.1). The airborne data consist of co-registered ALS and HSI data acquired from a Partenavia P.68-B Victor aircraft at an average altitude of approximately $1240~\mathrm{m}$ above ground level. HySpex VS-725 and Riegl VQ-780ii sensors were used to obtain HSI data with a ground sampling distance of $1~\mathrm{m}$ and ALS data with a density of $10~\mathrm{points}~\mathrm{m}^{-2}$, respectively. Airborne data were collected during one flight campaign on 4 September 2025. The surveyed area covers $68.58~\mathrm{km}^2$. The same processing workflow as described in Section~2.1 was used. Labels for a total of 512 reference trees were obtained from field measurements and assigned to 13 taxa classes, with 56 trees labelled as 'Other trees'. An additional 800 polygons were added as a `background’ class.
Visualisation of the reference area and training, validation, and test subsets used in experiments is presented in Figure~\ref{fig:study_area_bt}.

The division of the dataset into training/validation/test sets is presented in Table~\ref{tab:newds_ref_pixel_count}. As in the previous case, the split was tree-based, so pixels from the same tree were assigned to the same split. We used the same methodology as in the original experiment. Since the dataset was gathered during a different acquisition, there are minor differences in the ALS-derived features (221 vs 219 features in the original dataset). Parameters were optimised on the validation set. The neural network was trained for 500 epochs with a batch size of 512, the pseudo-label generation confidence threshold was 0.85, and the cohabitation scaling factor was set to 1.0. The remaining parameters were the same as in the original experiment.

\begin{table*}
\centering
\begin{tabular}{llrrrr}
\toprule
Tree taxa class & Class code & Total pixels & Train & Validation & Test \\
\midrule
\textit{Acer} spp. & Ace-spp & 644 & 243 & 172 & 229 \\
\textit{Alnus glutinosa} & Aln-glu & 2714 & 1829 & 507 & 378 \\
\textit{Betula} spp. & Bet-spp & 1637 & 967 & 173 & 497 \\
\textit{Fagus sylvatica} & Fag-syl & 281 & 167 & 46 & 68 \\
\textit{Larix} spp. & Lar-spp & 406 & 192 & 115 & 99 \\
\textit{Picea abies} & Pic-abi & 521 & 301 & 112 & 108 \\
\textit{Picea abies} - dead & Pic-dea & 181 & 107 & 50 & 24 \\
\textit{Pinus sylvestris} & Pin-syl & 3353 & 1837 & 903 & 613 \\
\textit{Pinus sylvestris} - dead & Pin-dea & 125 & 98 & 8 & 19 \\
\textit{Populus tremula} & Pop-tre & 86 & 81 & 2 & 3 \\
\textit{Quercus} spp. & Que-spp & 1559 & 623 & 498 & 438 \\
\textit{Salix} spp. & Sal-spp & 388 & 223 & 120 & 45 \\
\textit{Tilia} spp. & Til-spp & 190 & 140 & 48 & 2 \\
Other trees & Oth-tre & 681 & 364 & 171 & 146 \\
Background & Backgr & 18132 & 11581 & 3443 & 3108 \\
\bottomrule
\end{tabular}
\caption{Reference pixel counts by taxa class for the Bory Tucholskie dataset. }
\label{tab:newds_ref_pixel_count}
\end{table*}

\subsubsection{Cohabitation matrix verification}
\label{sec:bt_cohabitation_evaluation}
\begin{figure}
    \centering
    \begin{subfigure}{0.49\linewidth}
        \centering
        \includegraphics[width=\linewidth]{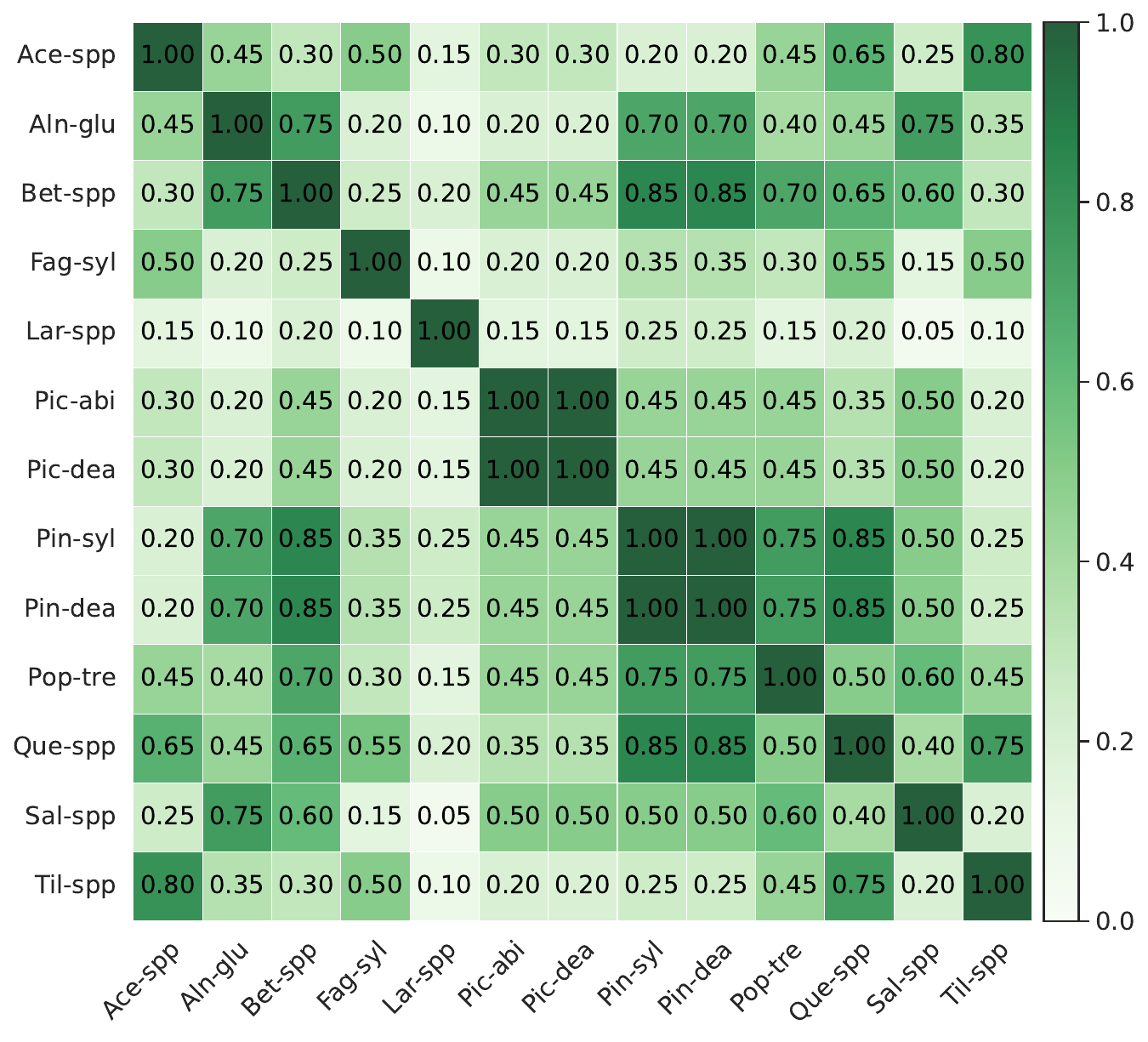}
        \caption{Cohabitation matrix}
        \label{fig:newds_cohabitation_matrix_a}
    \end{subfigure}
    \hfill
    \begin{subfigure}{0.49\linewidth}
        \centering
        \includegraphics[width=\linewidth]{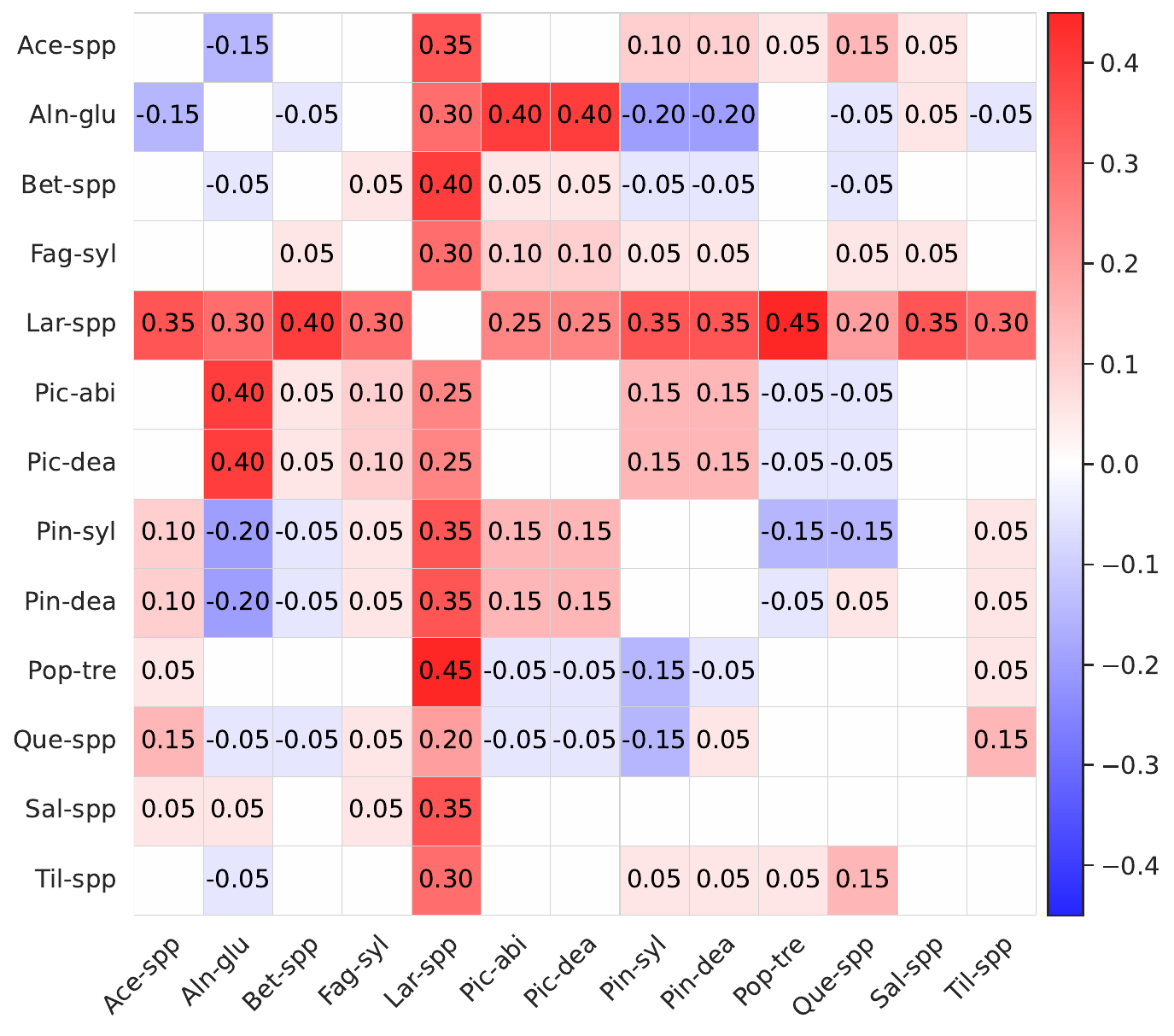}
        \caption{Expert updates}
        \label{fig:newds_cohabitation_matrix_b}
    \end{subfigure}
    \caption{LLM-based cohabitation matrix (a) and differences proposed by experts (b). Values are percentages, cells without value should be treated as zero. Bory Tucholskie dataset.}
    \label{fig:newds_cohabitation_matrix}
\end{figure}

The cohabitation matrix for the Bory Tucholskie dataset, along with its expert evaluation, is presented in Figure~\ref{fig:newds_cohabitation_matrix}. The values in the matrix are rounded to 0.05. The verification conducted by a botanist identified two significant errors in the LLM-based cohabitation matrix. The first concerns the ‘\textit{Larix}’ class, which includes the species \textit{Larix decidua} Mill. The LLM-derived cohabitation matrix assigned this species low cohabitation scores relative to most other taxa (0.05–0.25), reflecting its position outside the natural range supported by the literature-based habitat evidence. This is because the Bory Tucholskie is located outside the natural range of this species \cite{matras2008technical}, and its cohabitation with other species is, in normal circumstances, unlikely. In this particular case, however, the \textit{Larix decidua} species was artificially introduced into the area and occurs alongside other species; nevertheless, its locations are anthropogenic, and its coexistence with other species is the result of human intervention.  The situation is very similar to that of the second species, for which errors greater than average were also observed -- \textit{Picea abies}, spruce. Spruce is also not a natural component of the Bory Tucholskie ecosystem; it occurs outside its natural range~\cite{skroppa2003euforgen} there and was introduced by humans.

These two cases clearly illustrate the limitations of the LLM-based cohabitation matrix method. If it is based primarily on documented sources on species' habitats, it will be significantly less precise in areas where the species occur as a result of human intervention that puts them outside their natural habitats and therefore forces their cohabitation with species they would normally not coexist with.

\subsubsection{Results of verification}
Results of verification experiment on Bory Tucholskie dataset are presented in Table~\ref{tab:newds_class_metrics} and follow a similar pattern to those obtained for the Wigierski dataset (see Table~\ref{tab:class_metrics}). The strongest reference method was again MNF-IND-ALS, which achieved a macro-F1 of 72.19. The first-pass DSNN improved this result to 73.31 macro-F1, while the proposed DSNN+P reached the best overall result of 77.04 macro-F1. This corresponds to an improvement of 4.85 percentage points over MNF-IND-ALS and 3.73 percentage points over DSNN, confirming that the pseudo-labelled samples improve the neural classifier also in an independent study area.

The class-level results are generally consistent with this trend, although the best method varies between individual taxa. DSNN+P achieved the best results for several classes, including \textit{Larix} spp., \textit{Populus tremula}, \textit{Quercus} spp., \textit{Tilia} spp., and Background separation. DSNN alone was strongest for \textit{Alnus glutinosa}, \textit{Betula} spp., and \textit{Pinus sylvestris}. For classes represented by very few test pixels, such as \textit{Populus tremula} and \textit{Tilia} spp., the class-level results should be interpreted with caution. The aggregate results support the main finding from the primary dataset: combining HSI and ALS in the dual-stream network improves performance over the strongest classical reference pipeline, and pseudo-labelling provides an additional gain. The stands in the Bory Tucholskie and Wigierski National Parks differ in both species composition and the abundance of individual species. Consequently, the datasets differ in their legend classes and in the levels of accuracy achieved. These differences provide an additional, although limited, indication that the main conclusions may be transferable to other protected stands.

\begin{table*}
\centering
\begin{tabular}{llrrrrrr}
\toprule
Tree taxa class & Class code & HSI & MNF-IND & MNF-IND-ALS & DSNN & DSNN+P & Test pixels \\
\midrule
\textit{Acer} spp. & Ace-spp & 36.95 & 85.19 & \textbf{91.99} & 87.00 & 89.05 & 229 \\
\textit{Alnus glutinosa} & Aln-glu & 52.92 & 67.73 & 72.70 & \textbf{78.63} & 77.39 & 378 \\
\textit{Betula} spp. & Bet-spp & 81.56 & 94.86 & 96.03 & \textbf{97.65} & 96.83 & 497 \\
\textit{Fagus sylvatica} & Fag-syl & 40.06 & \textbf{66.50} & 56.65 & 57.11 & 61.66 & 68 \\
\textit{Larix} spp. & Lar-spp & 47.71 & 84.33 & 85.52 & 85.68 & \textbf{89.41} & 99 \\
\textit{Picea abies} & Pic-abi & 47.90 & \textbf{85.40} & 79.50 & 84.87 & 84.97 & 108 \\
\textit{Picea abies} - dead & Pic-dea & 47.34 & 66.29 & \textbf{75.89} & 69.62 & 72.70 & 24 \\
\textit{Pinus sylvestris} & Pin-syl & 83.97 & 94.34 & 93.70 & \textbf{96.68} & 96.49 & 613 \\
\textit{Pinus sylvestris} - dead & Pin-dea & 45.93 & 72.76 & \textbf{84.72} & 72.18 & 77.06 & 19 \\
\textit{Populus tremula} & Pop-tre & 29.59 & 18.77 & 7.02 & 33.27 & \textbf{66.76} & 3 \\
\textit{Quercus} spp. & Que-spp & 39.12 & 79.04 & 81.93 & 83.57 & \textbf{85.42} & 438 \\
\textit{Salix} spp. & Sal-spp & 15.16 & 52.37 & \textbf{82.54} & 79.64 & 67.21 & 45 \\
\textit{Tilia} spp. & Til-spp & 0.00 & 6.49 & 7.42 & 8.13 & \textbf{22.67} & 2 \\
Other trees & Oth-tre & 9.15 & 60.71 & 69.13 & 67.50 & \textbf{69.77} & 146 \\
Background & Backgr & 94.21 & 97.04 & 98.08 & 98.11 & \textbf{98.22} & 3108 \\
$\bar{x}$ & & 44.77 & 68.79 & 72.19 & 73.31 & \textbf{77.04} & \\
\bottomrule
\end{tabular}
\caption{Results comparison for tested methods on the second dataset -- Bory Tucholskie. Values are per-class F1 scores (reported as percentages, $\times 100$) and macro-averaged F1 ($\bar{x}$, bottom row). For the proposed Dual-Stream Neural Network (DSNN) and DSNN with Pseudo-labels (DSNN+P), results are averaged over five runs.}
\label{tab:newds_class_metrics}
\end{table*}

\subsection{Ablation study}
\label{sec:ablation_study}
We performed an ablation study to analyse the contribution of the pseudo-labelling procedure and the ecological cohabitation prior.  The results are reported in Tables~\ref{tab:ablation_Wig}, \ref{tab:ablation_BT}, and \ref{tab:ablation_reference}. In particular, we compared the proposed method with a diagonal-only variant, denoted as DSNN+D. In this setting, the cohabitation matrix has non-zero values only on the diagonal, so pseudo-label assignment is constrained to the same species class and does not use inter-species cohabitation information. Therefore, DSNN+D evaluates the effect of pseudo-label generation based on the treetop detector, the first-pass classifier, spatial proximity, and classifier confidence, but without the ecological cohabitation prior. The full DSNN+P variant additionally uses the cohabitation matrix, thereby incorporating ecological information on plausible species co-occurrence.

We also evaluated two more restrictive variants, denoted as DSNN(D) and DSNN(P), in which the network was trained only on pseudo-labelled pixels generated using the diagonal and cohabitation matrices, respectively. These experiments test whether the generated pseudo-labels contain sufficient class information to train a meaningful classifier without using the original labelled training samples in the second stage. Since supervised training requires at least one training example per class, these variants were evaluated only for classes represented in the corresponding pseudo-label set.

On the Wigierski dataset, the diagonal variant achieved a macro-F1 score of 78.57, which is between the first-pass DSNN result of 77.42 and the full DSNN+P result of 79.38. This indicates that the pseudo-labels generated without inter-species cohabitation information already improve the neural network, while the addition of the ecological cohabitation prior provides a further improvement. On the Bory Tucholskie dataset, DSNN achieved 73.31 macro-F1, while DSNN+D and DSNN+P achieved 76.99 and 77.04, respectively. In this case, the diagonal and cohabitation-based variants produced very similar results. For this dataset, the diagonal variant was additionally selected after a limited validation-based parameter optimisation; when evaluated with the same parameters as DSNN+P, it achieved a macro-F1 of 76.12. These results suggest that the proposed pseudo-label generation procedure consistently benefits the neural network, while the cohabitation prior either improves results or preserves comparable performance.

The pseudo-label-only variants, DSNN(D) and DSNN(P), achieved lower performance than models trained with both original labels and pseudo-labels. This is expected, since pseudo-labels are not intended to replace manually verified annotations, but rather to extend them. Nevertheless, the results are non-trivial, especially on the Bory Tucholskie dataset, where DSNN(D) and DSNN(P) achieved macro-F1 scores of 63.46 and 64.37, respectively. This indicates that the generated pseudo-labels contain meaningful class-discriminative information, although the best results are obtained when they are combined with the original labelled data.

Finally, Table~\ref{tab:ablation_reference} shows the effect of adding pseudo-labels to the reference methods. For the Wigierski dataset, the strongest reference method, MNF-IND-ALS, improved macro-F1 from 73.37 to 75.11, with small positive changes in overall accuracy ($+1.05$) and Kappa ($+1.51$).  For the Bory Tucholskie dataset, all three reference methods improved in macro-F1, with gains from $+0.09$ to $+3.86$ percentage points; overall accuracy and Kappa were also generally preserved or improved, with only a negligible $-0.01$ percentage-point change in overall accuracy for MNF-IND-ALS. This suggests that the pseudo-labels are not specific to the proposed neural architecture, although the improvement is more consistent for DSNN. A likely explanation is that the reference methods have more limited capacity to exploit the additional pseudo-labelled samples and may be more sensitive to residual pseudo-label noise.

As an additional assessment of the generated pseudo-labels, we compared them with the predictions of the first-pass classifier. Agreement was 94.4\% for Wigierski and 93.5\% for Bory Tucholskie, indicating that the pseudo-labelling procedure modifies only a relatively small fraction of the initial predictions. Direct comparison with available validation reference labels yielded no overlapping pseudo-labelled pixels for Wigierski and 17 pixels for Bory Tucholskie, corresponding to three spatially connected locations; all 17 agreed with the reference labels. The latter sample is too small to provide a reliable estimate of the overall pseudo-label error rate.

\begin{table*}
\centering
\begin{tabular}{llllllll}
\toprule
Tree taxa class & Class code & MNF-IND-ALS & DSNN(D) & DSNN(P) & DSNN & DSNN+D & DSNN+P \\
\midrule
Acer platanoides & Ace-pla & 70.80 & 73.37 & 43.69 & 73.76 & 76.46 & 75.25 \\
Alnus glutinosa & Aln-glu & 55.94 & 61.23 & 44.50 & 53.16 & 58.48 & 55.24 \\
Betula spp. & Bet-spp & 65.92 & 58.60 & 53.68 & 83.57 & 81.17 & 84.19 \\
Carpinus betulus & Car-bet & 10.47 & 18.91 & 1.64 & 26.10 & 27.37 & 22.73 \\
Fraxinus excelsior & Fra-exc & 74.19 & 17.83 & 0.00 & 80.67 & 86.34 & 83.65 \\
Larix decidua & Lar-dec & 87.21 & 74.04 & 45.41 & 87.40 & 87.98 & 87.06 \\
Picea abies & Pic-abi & 91.06 & 87.21 & 80.97 & 91.53 & 92.44 & 92.22 \\
Picea abies - dead & Pic-dea & 97.52 & 90.59 & 90.52 & 96.21 & 97.23 & 97.35 \\
Pinus sylvestris & Pin-syl & 76.53 & 83.05 & 74.98 & 80.81 & 82.08 & 88.89 \\
Pinus sylvestris - dead & Pin-dea & 86.11 & 13.89 & 0.00 & 81.13 & 77.98 & 80.80 \\
Populus tremula & Pop-tre & 60.54 & 71.23 & 50.96 & 68.38 & 63.13 & 67.13 \\
Quercus robur & Que-rob & 86.85 & 92.66 & 91.41 & 90.14 & 92.54 & 91.30 \\
Salix trees & Sal-tre & 56.53 & 76.71 & 56.67 & 85.77 & 87.68 & 86.36 \\
Salix bushes & Sal-bus & 86.90 & 70.74 & 26.81 & 81.11 & 82.44 & 80.95 \\
Tilia cordata & Til-cor & 89.26 & 87.41 & 89.13 & 90.55 & 91.73 & 91.96 \\
Deciduous trees - dead & Dec-dea & 60.00 & 1.82 & 7.98 & 68.67 & 60.32 & 75.43 \\
Other trees & Oth-tre & 65.54 & -- & -- & 68.35 & 69.75 & 68.98 \\
Background & Backgr & 99.22 & -- & -- & 99.05 & 99.10 & 99.11 \\
$\bar{x}$ &  & 73.37 & 61.21 & 47.40 & 77.42 & 78.57 & 79.38 \\
\bottomrule
\end{tabular}
\caption{Results of the ablation study for the Wigierski dataset (F1 macro). MNF-IND-ALS is the best reference method. DSNN(D) and DSNN(P) are models trained on pseudo-labels generated using the diagonal and cohabitation matrices, respectively. DSNN denotes the results of the first-pass NN training. DSNN+D uses pseudo-labels generated with the diagonal matrix, i.e., without cohabitation, whereas DSNN+P denotes the proposed method}
\label{tab:ablation_Wig}
\end{table*}

\begin{table*}
\centering
\begin{tabular}{llrrrrrr}
\toprule
Tree taxa class & Class code & MNF-IND-ALS & DSNN(D) & DSNN (P) & DSNN & DSNN+D & DSNN+P \\
\midrule
\textit{Acer} spp. & Ace-spp & 91.99 & 81.17 & 59.75 & 87.00 & 89.51 & 89.05 \\
\textit{Alnus glutinosa} & Aln-glu & 72.70 & 80.99 & 72.11 & 78.63 & 78.23 & 77.39 \\
\textit{Betula} spp. & Bet-spp & 96.03 & 96.98 & 94.07 & 97.65 & 97.52 & 96.83 \\
\textit{Fagus sylvatica} & Fag-syl & 56.65 & 65.21 & 55.06 & 57.11 & 69.26 & 61.66 \\
\textit{Larix} spp. & Lar-spp & 85.52 & 83.35 & 84.15 & 85.68 & 92.70 & 89.41 \\
\textit{Picea abies} & Pic-abi & 79.50 & 81.86 & 73.26 & 84.87 & 90.07 & 84.97 \\
\textit{Picea abies} - dead & Pic-dea & 75.89 & 74.79 & 61.70 & 69.62 & 72.11 & 72.70 \\
\textit{Pinus sylvestris} & Pin-syl & 93.70 & 96.50 & 92.57 & 96.68 & 97.21 & 96.49 \\
\textit{Pinus sylvestris} - dead & Pin-dea & 84.72 & 65.01 & 77.34 & 72.18 & 78.40 & 77.06 \\
\textit{Populus tremula} & Pop-tre & 7.02 & 10.00 & 0.00 & 33.27 & 50.55 & 66.76 \\
\textit{Quercus} spp. & Que-spp & 81.93 & 75.17 & 65.13 & 83.57 & 86.30 & 85.42 \\
\textit{Salix} spp. & Sal-spp & 82.54 & 13.92 & 37.34 & 79.64 & 70.59 & 67.21 \\
\textit{Tilia} spp. & Til-spp & 7.42 & 0.00 & -- & 8.13 & 13.71 & 22.67 \\
Other trees & Oth-tre & 69.13 & -- & -- & 67.50 & 70.51 & 69.77 \\
Background & Backgr & 98.08 & -- & -- & 98.11 & 98.24 & 98.22 \\
$\bar{x}$ & & 72.19 & 63.46 & 64.37 & 73.31 & 76.99 & 77.04 \\
\bottomrule
\end{tabular}
\caption{Results of the ablation study for the Bory Tucholskie dataset (F1 macro). MNF-IND-ALS is the best reference method. DSNN(D) and DSNN(P) are models trained on pseudo-labels generated using the diagonal and cohabitation matrices, respectively. DSNN denotes the results of the first-pass NN training. DSNN+D uses pseudo-labels generated with the diagonal matrix, i.e., without cohabitation, whereas DSNN+P denotes the proposed method}
\label{tab:ablation_BT}
\end{table*}

\begin{table*}
\centering
\begin{tabular}{lllrrr}
\toprule
Dataset & Metric & Method & Original & Pseudolabels & Difference \\
\midrule
\multirow{9}{*}{Wigierski} & \multirow{3}{*}{F1 macro} & HSI & 42.54 & 44.11 & +1.57 \\
 & & MNF-IND & 71.03 & 71.74 & +0.71 \\
 & & MNF-IND-ALS & 73.37 & 75.11 & +1.74 \\
 & \multirow{3}{*}{Overall accuracy} & HSI & 74.64 & 74.81 & +0.17 \\
 & & MNF-IND & 86.37 & 87.11 & +0.74 \\
 & & MNF-IND-ALS & 87.85 & 88.90 & +1.05 \\
 & \multirow{3}{*}{Kappa} & HSI & 63.11 & 63.61 & +0.50 \\
 & & MNF-IND & 80.42 & 81.51 & +1.09 \\
 & & MNF-IND-ALS & 82.68 & 84.19 & +1.51 \\
\midrule
\multirow{9}{*}{Bory Tucholskie} & \multirow{3}{*}{F1 macro} & HSI & 44.77 & 46.44 & +1.67 \\
 & & MNF-IND & 68.79 & 72.65 & +3.86 \\
 & & MNF-IND-ALS & 72.19 & 72.28 & +0.09 \\
 & \multirow{3}{*}{Overall accuracy} & HSI & 78.89 & 79.75 & +0.86 \\
 & & MNF-IND & 90.20 & 90.73 & +0.53 \\
 & & MNF-IND-ALS & 91.71 & 91.70 & -0.01 \\
 & \multirow{3}{*}{Kappa} & HSI & 68.38 & 70.21 & +1.83 \\
 & & MNF-IND & 85.57 & 86.39 & +0.82 \\
 & & MNF-IND-ALS & 87.89 & 87.90 & +0.01 \\
\bottomrule
\end{tabular}
\caption{Ablation study -- comparison of reference methods without (original) and with pseudolabels.}
\label{tab:ablation_reference}
\end{table*}

\section{Conclusions}
Despite the complexity of hyperspectral tree species classification with sparse labels, the proposed dual-stream architecture achieved high accuracy and outperformed strong classical HSI and HSI–ALS reference pipelines. The proposed lightweight dual-stream architecture achieved good performance. Our results also demonstrate that LLM-assisted priors can provide useful initial co-occurrence estimates, but should be treated as approximate ecological hypotheses requiring expert verification. Since obtaining such priors is costly and typically requires expert knowledge, the ability to estimate them reliably represents a promising step toward developing expert algorithms that combine machine learning with the estimation of biological parameters. The proposed pseudo-labelling approach leverages cohabitation to extend the training set and improve model accuracy. Overall, our results show that integrating neural network architectures with ecological priors and semi-supervised learning can significantly improve tree species classification under limited-label conditions.

\section{Acknowledgements}
The authors acknowledge the use of AI-based tools, such as ChatGPT and Grammarly, for assistance in editing, grammar enhancement, and spelling checks during the preparation of this manuscript.

Funding: 
this work was partially supported by project IITIS/BW/08/25 `Multi-modal neural networks for tree species classification from remote sensing data', financed by the Institute of Theoretical and Applied Informatics, Polish Academy of Sciences. 
Ground reference data and a complete set of airborne data were obtained as part of two projects: (1) `Acquisition of multi-source remote sensing data and their analysis for the area of Wigry National Park with a part of Wigry Lake and the Czarna Hańcza River' project, financed by the Operational Programme Infrastructure and Environment under programme 2.4.4d – assessment of the state of natural resources in national parks using modern remote sensing technologies; and (2) `Preparation of sectoral studies and a draft protection plan for Bory Tucholskie National Park' project, carried out under the European Funds for Infrastructure, Climate and Environment 2021–2027 programme, Measure FENX.01.05.2.

\appendix
\section{Appendix}

\subsection{Cohabitation matrix prompt}
\label{sec:cohabitation_prompt}
A prompt to generate the cohabitation matrix by LLM system:
\begin{prompt}
You are a meticulous forest ecology expert and digital librarian.
You perform careful web research and return clean, verifiable 
outputs. If you have a claim you always provide the sources to 
back it up 
    
# TASK 
1) Normalize species_list: 
- Trim lines; drop empties/duplicates. 
- Exclude non-species lines (e.g., "background", "inne", "lis_mar"). 
- If there is a ambiguity in understanding of the name (e.g. it can mean two different species) 
use all options as distinct species. 
- If a line contains “spp.” mark flags.genus_level=true when used in pairs. 
2) Generate ALL unordered species pairs (A,B), where A != B, A <= B alphabetically within the pair. 
Consider neighbors at stand-scale (same stand or within {distance_m} m). 
3) Divide those pairs into subgroups of 10-12. Prepare report for each subgroup 
and in the end join them. 
4) For EACH pair: 
- Use web.search() to retrieve {min_sources_per_pair}–{max_sources_per_pair} authoritative, 
verifiable sources that speak to stand-scale co- occurrence or shared habitat in the given 
region. USING web.search() IS CRUCIAL DO NOT SKIP IT. 
- For each species provide a source of its natural habitats. USE THEM in your reports, so 
each pair has at least two sources - natural habitats of A and natural habitats of B - The 
next sources if present should be about specific co-occurences. 
- Take into consideration {region} and {additional environmental information}
- Prioritize: 
a) EUNIS/JNCC/EEA official habitat pages, 
b) Publisher DOI landing pages (via doi.org), 
c) Government/NGO vegetation surveys or credible databases. 
- If only weaker sources exist, BE SURE to include them (with working URLs), set 
confidence="low" or "medium", and briefly explain in rationale why evidence is tentative. 
- Those weaker sources may include university pages, or lesser known books, articles and 
reports that are accessible. 
- Be sure that if you find ANY source (be it prioritized sources or the weaker sources) for 
given pair this pair MUST NOT be left without sources. 
- If out-of-region (e.g., North American taxa with European species), set 
flags.out_of_region=true and lower the score appropriately. 
4) Scoring guidance: 
- score in [0,1] is the likelihood of stand-scale adjacency in {region}. 
- Provide score_low and score_high as a plausible range. 
- High scores (>=0.7) only with strong, in-region, stand-scale evidence. 0.5 = plausible 
but mixed evidence; 0.0–0.2 = unlikely or habitat mismatch. 
- Keep rationale <= 240 chars; no newlines. 
- If you have weaker sources set confidence to low. 
5) If you cannot find sources, for neither cohabitation nor habitats do not include the pair. 
6) Citations quality and formatting: 
- DO NOT fabricate DOIs. If known, put bare DOI (e.g., "10.1000/xyz123") in "doi"; also include 
the working URL in "url". 
- Each source MUST have at least a working URL or a DOI landing page. 
- Remove any bracket markers (no `$[\ldots]$'), no smart quotes, no markdown. 
- No newlines in any JSON string. 
7) Output requirements: 
- Output EXACTLY TWO sections in order, separated by these plain-text markers: 

===CSV=== 

cohabitation symmetric matrix with rows and columns from all the tree_species with 
cohabitation scores from the initial tree_species, -1.0 when score is unavailable and 1.0 
on the diagonal (self-cohabitation), even if you did not find any cohabitation for them. 

===LATEX=== <standalone .tex document> 
- The LaTeX must: 
* Use article class; packages: geometry, hyperref, longtable, booktabs, url, inputenc/fontenc. 
* Include a longtable with columns: Species A, Species B, Score, Range, Confidence. 
* Longtable rows must be separated with double "\\" char. 
* After the table, include per-pair subsections with: italic Rationale, then bullet-style 
paragraphs listing sources (Title — Authors; Year; DOI/URL). Escape LaTeX special chars. 
* Show the score and the (low–high) range in each subsection header. 
* For each pair you should have a subsection naming 
- species A, species B 
- Habitat A <citation> - Habitat B <citation> 
- Other sources of co-occurence or lack of thereof (use citations) 
- Rationale for your score based on Habitat A, Habitat B and other Sources - use bibliography
and citations instead of inline citing. 
\end{prompt}

\printcredits

\bibliographystyle{unsrt}

\bibliography{bibliography}


\begin{center}
\small
\onecolumn
\begin{longtable}{>{\raggedright\arraybackslash}p{3.5cm} >{\raggedright\arraybackslash}p{11.0cm}}
\caption{Hyperspectral-derived indices used in the study.\label{tab:hs_ind}}\\
\hline
\textbf{Index short name} & \textbf{Full name} \\
\hline
\endfirsthead

\multicolumn{2}{l}{\small\itshape continued from previous page} \\
\hline
\textbf{Short\_Name} & \textbf{Meaning} \\
\hline
\endhead

\hline
\multicolumn{2}{r}{\small\itshape continued on next page} \\
\endfoot

\hline
\endlastfoot

ALUI & Alunite Index \\
ARI1 & Anthocyanin Reflectance Index 1 \\
ARI2 & Anthocyanin Reflectance Index 2 \\
ARVI & Atmospherically Resistant Vegetation Index \\
BAI & Burn Area Index \\
CRI1 & Carotenoid Reflectance Index 1 \\
CRI2 & Carotenoid Reflectance Index 2 \\
CAI & Cellulose Absorption Index \\
CLAI & Clay Alteration Index \\
CLAY & Clay Minerals Ratio \\
DVI & Difference Vegetation Index \\
DWSI & Disease Water Stress Index \\
DOLI & Dolomite Index \\
EVI & Enhanced Vegetation Index \\
FEAI & Ferric Iron Alteration Index \\
Fe2+ & Ferrous Iron (Fe2+) Index \\
Ferrous & Ferrous Minerals Ratio \\
FESI & Ferrous Silicates Index \\
GEMI & Global Environmental Monitoring Index \\
GARI & Green Atmospherically Resistant Index \\
GCI & Green Chlorophyll Index \\
GDVI & Green Difference Vegetation Index \\
GLI & Green Leaf Index \\
GNDVI & Green Normalized Difference Vegetation Index \\
GOSAVI & Green Optimized Soil Adjusted Vegetation Index \\
GRVI & Green Ratio Vegetation Index \\
GSAVI & Green Soil Adjusted Vegetation Index \\
GVI & Green Vegetation Index \\
OH1 & Hydroxyl-Bearing (OH) Altered Minerals Index 1 \\
OH2 & Hydroxyl-Bearing (OH) Altered Minerals Index 2 \\
OH3 & Hydroxyl-Bearing (OH) Altered Minerals Index 3 \\
IPVI & Infrared Percentage Vegetation Index \\
Iron Oxide & Iron Oxide Ratio \\
KAI1 & Kaolinite Index 1 \\
KAI2 & Kaolinite Index 2 \\
KAI3 & Kaolinite Index 3 \\
LATI & Laterite Index \\
LAI & Leaf Area Index \\
LCI & Leaf Chlorophyll Index \\
LWVI1 & Leaf Water Vegetation Index 1 \\
LWVI2 & Leaf Water Vegetation Index 2 \\
LCAI & Lignin Cellulose Absorption Index \\
METHANE & Methane Index \\
MCARI & Modified Chlorophyll Absorption Ratio Index \\
MCARI2 & Modified Chlorophyll Absorption Radio Index Improved \\
MNLI & Modified Non-Linear Index \\
MNDWI & Modified Normalized Difference Water Index \\
MRENDVI & Modified Red Edge Normalized Difference Vegetation Index \\
MRESR & Modified Red Edge Simple Ratio \\
MSR & Modified Simple Ratio \\
MSAVI2 & Modified Soil Adjusted Vegetation Index 2 \\
MTVI & Modified Triangular Vegetation Index \\
MTVI2 & Modified Triangular Vegetation Index - Improved \\
MSI & Moisture Stress Index \\
MONI & Montmorillonite Index \\
MUSI & Muscovite Index \\
NLI & Non-Linear Index \\
NBR & Normalized Burn Ratio \\
NDBI & Normalized Difference Built-Up Index \\
NDII & Normalized Difference Infrared Index \\
NDLI & Normalized Difference Lignin Index \\
NDMI & Normalized Difference Mud Index \\
NDNI & Normalized Difference Nitrogen Index \\
NDSI & Normalized Difference Snow Index \\
NDVI & Normalized Difference Vegetation Index \\
NDWI & Normalized Difference Water Index \\
NMDI & Normalized Multiband Drought Index \\
NPCI & Normalized Pigment Chlorophyll Index \\
OSAVI & Optimized Soil Adjusted Vegetation Index \\
PHEI & Phengitic Index \\
PRI & Photochemical Reflectance Index \\
PHAI & Phyllic Alteration Index \\
PSRI & Plant Senescence Reflectance Index \\
RENDVI & Red Edge Normalized Difference Vegetation Index \\
REPI & Red Edge Position Index \\
RGRI & Red Green Ratio Index \\
RDVI & Renormalized Difference Vegetation Index \\
SR & Simple Ratio \\
SAVI & Soil Adjusted Vegetation Index \\
SIPI1 & Structure Independent Pigment Index \\
SIPI & Structure Insensitive Pigment Index \\
SGI & Sum Green Index \\
TCARI & Transformed Chlorophyll Absorption Reflectance Index \\
TDVI & Transformed Difference Vegetation Index \\
TGI & Triangular Greenness \\
TVI & Triangular Vegetation Index \\
VARI & Visible Atmospherically Resistant Index \\
VREI1 & Vogelmann Red Edge Index 1 \\
WBI & Water Band Index \\
WDRVI & Wide Dynamic Range Vegetation Index \\
WV-BI & WorldView Built-Up Index \\
WV-VI & WorldView Improved Vegetative Index \\
WV-II & WorldView New Iron Index \\
WV-NHFD & WorldView Non-Homogeneous Feature Difference \\
WV-SI & WorldView Soil Index \\
WV-WI & WorldView Water Index \\

\end{longtable}
\end{center}

\vspace{0.5em}
\noindent\footnotesize
\textit{Note:} The indices listed in Table~\ref{tab:hs_ind} are all calculated using ENVI Spectral Indices. Formulas and sources are available in the ENVI on-line manual \cite{ENVIApiGuide}.

\begin{center}
\small
\onecolumn
\begin{longtable}{>{\raggedright\arraybackslash}p{3.5cm} >{\raggedright\arraybackslash}p{11.0cm}}
\caption{LiDAR-derived structural metrics used in the study.\label{tab:lidar_metrics}}\\
\hline
\textbf{Short\_Name} & \textbf{Meaning} \\
\hline
\endfirsthead

\multicolumn{2}{l}{\small\itshape continued from previous page} \\
\hline
\textbf{Short\_Name} & \textbf{Meaning (English)} \\
\hline
\endhead

\hline
\multicolumn{2}{r}{\small\itshape continued on next page} \\
\endfoot

\hline
\endlastfoot

CC & Canopy Cover — Ratio of the number of first returns originating from vegetation classes to the number of first returns from all classes. \\
gap\_fraction & Gap fraction — Ratio of returns classified as ground (all returns below a given threshold [2 m] are treated as ground) to all returns. \\
LAI & Leaf Area Index estimate — area of leaf surface per unit ground area, expressing canopy leaf density from a nadir perspective. \\
herbaceous\_density & Ratio of returns from low and medium vegetation classes below 3 m to all returns within a pixel. \\
shrub\_density & Ratio of returns from vegetation in the 3–6 m height range to all returns within a pixel. \\
tree\_density & Ratio of returns from vegetation taller than 6 m to all returns within a pixel. \\
deviation\_covar & Covariance of pulse-shape deviation values within a grid cell. \\
deviation\_max & Maximum pulse-shape deviation within a grid cell. \\
deviation\_mean & Mean pulse-shape deviation within a grid cell. \\
deviation\_median & Median pulse-shape deviation within a grid cell. \\
deviation\_min & Minimum pulse-shape deviation within a grid cell. \\
deviation\_range & Range (maximum minus minimum) of pulse-shape deviation values within a grid cell. \\
deviation\_rms & Root mean square of pulse-shape deviation values within a grid cell. \\
deviation\_var & Variance of pulse-shape deviation values within a grid cell. \\
dq\_25 & 25th percentile of pulse-shape deviation values within a grid cell. \\
dq\_75 & 75th percentile of pulse-shape deviation values within a grid cell. \\
iqr & Interquartile range of pulse-shape deviation values within a grid cell. \\
FHD & Foliage Height Diversity — vertical distribution of foliage quantified as the proportional horizontal foliage cover per height layer relative to total cover. \\
FHDGT & Foliage Height Diversity computed for heights above 1 m, emphasizing taller vegetation layers. \\
bin\_0.0\_0.5 & Number of returns in the height interval 0.0–0.5 m. \\
bin\_0.5\_1.0 & Number of returns in the height interval 0.5–1.0 m. \\
bin\_1.0\_1.5 & Number of returns in the height interval 1.0–1.5 m. \\
bin\_1.5\_2.0 & Number of returns in the height interval 1.5–2.0 m. \\
bin\_2.0\_5.0 & Number of returns in the height interval 2–5 m. \\
bin\_5.0\_10.0 & Number of returns in the height interval 5–10 m. \\
bin\_10.0\_15.0 & Number of returns in the height interval 10–15 m. \\
bin\_15.0\_20.0 & Number of returns in the height interval 15–20 m. \\
bin\_20.0\_25.0 & Number of returns in the height interval 20–25 m. \\
bin\_25.0\_30.0 & Number of returns in the height interval 25–30 m. \\
bin\_30.0 & Number of returns above 30 m. \\
CC\_over2m & Canopy cover computed using first returns from vegetation above 2 m. \\
H99-H50 & Difference between the 99th and 50th percentiles of height distribution. \\
H99-H25 & Difference between the 99th and 25th percentiles of height distribution. \\
H90-H50 & Difference between the 90th and 50th percentiles of height distribution. \\
H90-H25 & Difference between the 90th and 25th percentiles of height distribution. \\
HCV & Coefficient of variation of heights derived from first returns. \\
hL1 & First L-moment of the height distribution. \\
hL2 & Second L-moment of the height distribution. \\
hL3 & Third L-moment of the height distribution. \\
hL4 & Fourth L-moment of the height distribution. \\
hLskw & L-skewness of the height distribution. \\
hLkurt & L-kurtosis of the height distribution. \\
zmean & Mean height value. \\
zmedian & Median height value. \\
zmax & Maximum height value. \\
zsd & Standard deviation of height values. \\
zskew & Skewness of the height distribution. \\
zentropy & Shannon entropy of the height distribution histogram within a grid cell. \\

\end{longtable}
\end{center}

\vspace{0.5em}
\noindent\footnotesize
\textit{Note:} The metrics listed in Table~\ref{tab:lidar_metrics} constitute a compilation of original (custom-developed) metrics and established LiDAR-derived metrics adapted from the \texttt{lidR} framework and from published studies on L-moment-based characterization of forest vertical structure \cite{lidRbook,valbuena_lmoments_2017}.
\twocolumn

\end{document}